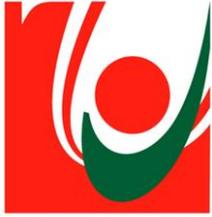
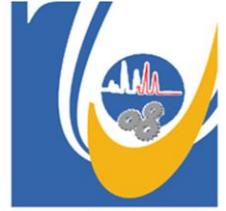

# LEBANESE UNIVERSITY
## DOCTORAL SCHOOL OF SCIENCES AND TECHNOLOGY

**Master Thesis**

Submitted in Partial Fulfillment of the
requirements for the Degree of
Master2 Research
in
Signal, Telecoms, Image and Speech (STIP)

"Super Resolution image reconstruct via total variation-based image deconvolution: a majorization-minimization approach"

Presented by

Mouhamad Houssien Chehaitly

Defended on:   7-9-2011         before the jury:

| | |
|---|---|
| **Dr.** Rima Hleiss | Supervisor |
| **Dr.** Ali Beydoun | Reporter |
| **Pr.** Yasser Mohanna | Examiner |

# Table of Contents

# Abstract

The objective of this work is the Super Resolution reconstruct of image sequences with Total Variation regulazer. We consider in particular images of scenes for which the point-to-point image transformation is a plane projective transformation.

We first describe the imaging observation model, an interpolation and Fusion estimator, and Projection on Convex Sets (POCS) of the super-resolution image.
We explain the motion and compute the optical flow of a sequence of images. In fact, we use the Horn-Shunck algorithm to estimate motion.
We then propose a Total Variation regulazer via Majorization-Minimization approach to obtain a suitable result. Super Resolution restoration from motion measurements is also discussed. Finally a simulations part demonstrates the power of the proposed methodology.

As expected, this model does not give results in real-time, as one can see in the numerical experiments section, but is the cornerstone for the future approaches.

# Bibliography

List of abbreviations:

HR    High Resolution
CCD   Charge-Coupled Device
LR     Low Resolution
SR     Super Resolution
ROI    Region Of Interest
DVR   Digital Video Recorder
CT     Computed Tomography
MRI   Magnetic Resonance Imaging
PSF   Point Spread Function
POCS   Projection Onto Convex Sets
TV     Total Variation
MM   Majorization-Minimization
CG    Conjugate Gradient
EM    Expectation-Maximization

Figures:

# Chapter I

# Introduction

## I.1 – Historic view

In most electronic imaging applications, images with high resolution (HR) are desired and often required. HR means that pixel density within an image is high, and the performance of many applications can be improved if an HR image is provided.

Since the 1970s, charge-coupled device (CCD) and CMOS image sensors have been widely used to capture digital images. Although these sensors are suitable for most imaging applications, the current resolution level and consumer price will not satisfy the future demand. For example, people want an inexpensive HR digital camera/camcorder or see the price gradually reduce, and scientists often need a very HR level close to that of an analog 35mm film that has no visible artifacts when an image is magnified. Thus, finding a way to increase the current resolution level is needed.

The most direct solution to increase spatial resolution is to reduce the pixel size (i.e., increase the number of pixels per unit area) by sensor manufacturing techniques. As the pixel size decreases, however, the amount of light available also decreases. It generates shot noise that degrades the image quality severely. To reduce the pixel size without suffering the effects of shot noise, therefore, there exists the limitation of the pixel size reduction, and the optimally limited pixel size is estimated at about 40 $\mu m^2$ for a 0.35 $\mu m^2$ for 0.14 $nm^2$ CMOS process. The current image sensor technology has almost reached this level.

Another approach for enhancing the spatial resolution is to increase the chip size, which leads to an increase in capacitance [1]. Since large capacitance makes it difficult to speed up a charge transfer rate, this approach is not considered effective. The high cost for high precision optics and image sensors is also an important concern in many commercial applications regarding HR imaging.

Therefore, a new approach toward increasing spatial resolution is required to overcome these limitations of the sensors and optics manufacturing technology.

One promising approach is to use signal processing techniques to obtain an HR image (or sequence) from observed multiple low resolution (LR) images. Recently, such a resolution enhancement approach has been one of the most active research areas, and it is called super resolution (SR) (or HR) image reconstruction or simply resolution enhancement in the literature [1]-[2].

The major advantage of the signal processing approach is that it may cost less and the existing LR imaging systems can be still utilized.

## I.2 - How can we obtain an HR image from multiple LR images?

The basic premise for increasing the spatial resolution in SR techniques is the availability of multiple LR images captured from the same scene.

In SR, typically, the LR images represent different "looks" of the same scene. That is, LR images are subsampled (aliased) as well as shifted with subpixel precision. If the LR images are shifted by integer units, then each image contains the same information, and thus there is no new information that can be used to reconstruct an HR image. If the LR images have different subpixel shifts from each other and if aliasing is present, however, then each image cannot be obtained from the others.

In this case, the new information contained in each LR image can be exploited to obtain an HR image. To obtain different looks at the same scene, some relative scene motions must exist from frame to frame via multiple scenes or video sequences. Multiple scenes can be obtained from one camera with several captures or from multiple cameras located in different positions. These scene motions can occur due to the controlled motions in imaging systems, (e.g., images acquired from orbiting satellites).

The same is true of uncontrolled motions, (e.g., movement of local objects or vibrating imaging systems).If these scene motions are known or can be estimated within subpixel accuracy and if we combine these LR images, SR image reconstruction is possible as illustrated in Figure I.1.

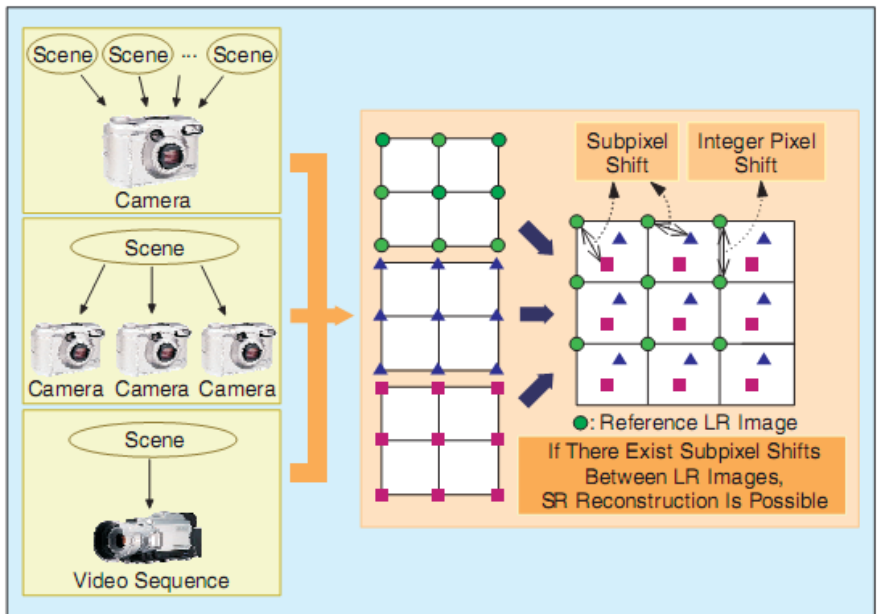

*Figure I.1 - Basic premise for Super-Resolution*

In the process of recording a digital image, there is a natural loss of spatial resolution caused by the optical distortions (out of focus, diffraction limit, etc.), motion blur due to limited shutter speed, noise that occurs within the sensor or during transmission, and insufficient sensor density as shown in Figure I.2.

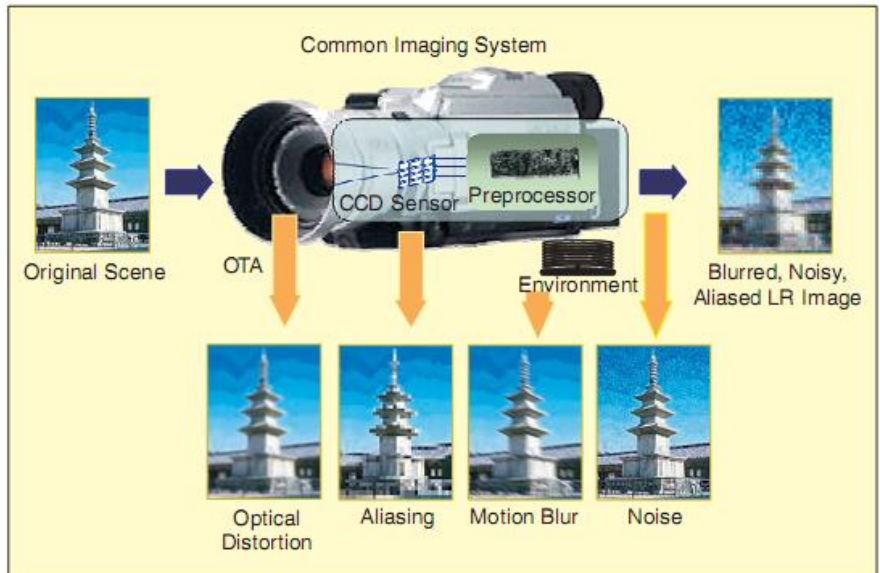

*Figure I.2 - Common image acquisition system.*

Thus, the recorded image usually suffers from blur, noise, and aliasing effects. Al though the main concern of an SR algorithm is to reconstruct HR images from undersampled LR images; it covers image restoration techniques that produce high quality images from noisy, blurred images.

A related problem to SR techniques is image restoration, which is a well-established area in image processing applications [3]-[4].

The goal of image restoration is to recover a degraded (e.g., blurred, noisy) image, but it does not change the size of image. In fact, restoration and SR reconstruction are closely related theoretically, and SR reconstruction can be considered as a second-generation problem of image restoration.

Another problem related to SR reconstruction is image interpolation that has been used to increase the size of a single image. Although this field has been extensively studied [5]-[6], the quality of an image magnified from an aliased LR image is inherently limited even though the ideal sinc basis function is employed. That is, single image interpolation cannot recover the high-frequency components lost or degraded during the LR sampling process. For this reason, image interpolation methods are not considered as SR techniques.

To achieve further improvements in this field, the next step requires the utilization of multiple data sets in which additional data constraints from several observations of the same scene can be used. The fusion of information from various observations of the same scene allows SR reconstruction of the scene.

So, we can simplify easily the Super-Resolution in the three steps: image registration, fusion, and deblurring [7]

## I.3 – Applications

The SR image reconstruction is proved to be useful in many practical cases where multiple frames of the same scene can be obtained, including medical imaging, satellite imaging, and video applications. One application is to reconstruct a higher quality digital image from LR images obtained with an inexpensive LR camera/camcorder for printing or frame freeze purposes.

Typically, with a camcorder, it is also possible to display enlarged frames successively.

Synthetic zooming of region of interest (ROI) is another important application in surveillance, forensic, scientific, medical, and satellite imaging. For surveillance or forensic purposes, a digital video recorder (DVR) is currently replacing the CCTV system, and it is often needed to magnify objects in the scene such as the face of a criminal or the license plate of a car. The SR technique is also useful in medical imaging such as computed tomography (CT) and magnetic resonance imaging (MRI) since the acquisition of multiple images is possible while the resolution quality is limited. In satellite imaging applications such as remote sensing and LANDSAT, several images of the same area are usually provided, and the SR technique to improve the resolution of target can be considered. Another application is conversion from an NTSC video signal to an HDTV signal since there is a clear and present need to display a SDTV signal on the HDTV without visual artifacts. And not finally the performance of pattern recognition in computer vision can be obtained if an HR image is provided.

The goal of this work is to introduce the concept of motion in a SR algorithm and to provide a review regularize terms.

To this purpose, I divide my work in four chapters:
The first one is a historic view and introduction of SR. In the second one, we descript in details the model to obtain, and we present the technical review of various existing SR methodologies which are often employed. Because the importance of motion, we specify the chapter 3 to present and explain the motion estimation with optical flow ,and to introduce the existing motion in frame of image SR. In the last chapter, we use a Regularization to adjust a Super Resolution Reconstruction in the case of single image observation and in the general case of a sequence of images.

# Chapter II

# Super Resolution

## II.1 – Introduction

In this chapter, we present an approach toward the SR restoration problem. Simplicity and direct connection to the problem of single image restoration (from one measured image) are the main benefits of this approach. Thus, the various known methods to restore one image from one measured image are easily generalized to general problem of single image restoration from several measured images, that is video sequences usually contain a large overlap between successive frames, and regions in the scene are sampled in several images. This multiple sampling can be used to achieve images with a higher spatial resolution. The process of reconstructing a HR image from several images covering the same region in the world is called SR. Additional tasks may include the reconstruction of a HR video sequence, or a HR 3D model of the scene.

A common model for SR is present by the following.

The LR input images are the result of projection of a HR image onto the image plane, followed by sampling. The goal is to find the HR image which fits this model.

In mathematical language given K images $\{Y^{(n)}\}$ (n=1, 2, ... , K) of size M1*M2 , we aim to find the image X of size N1*N2 , which minimizes the error function:

$$e(X) = \sum_{n=1}^{K} ||P_n(X) - Y^{(n)}||^2 \qquad (1)$$

where:
1. $||.||$ - Can be any norm, usually $l^2$.
2. $P_n(X)$ is the projection of X onto the coordinate system and sampling grid of image $Y^{(n)}$.

When this optimization problem does not have a single solution, additional constraints may be added, expressing prior assumptions on X, such as smoothness.

The projection $P_n(X)$ is usually modeled by four stages:

a. Geometric Transformation (Motion).

b. Blurring.
c. Decimation or Subsampling.
d. Additive Noise.

We can show the effect for every step in figure II.1:

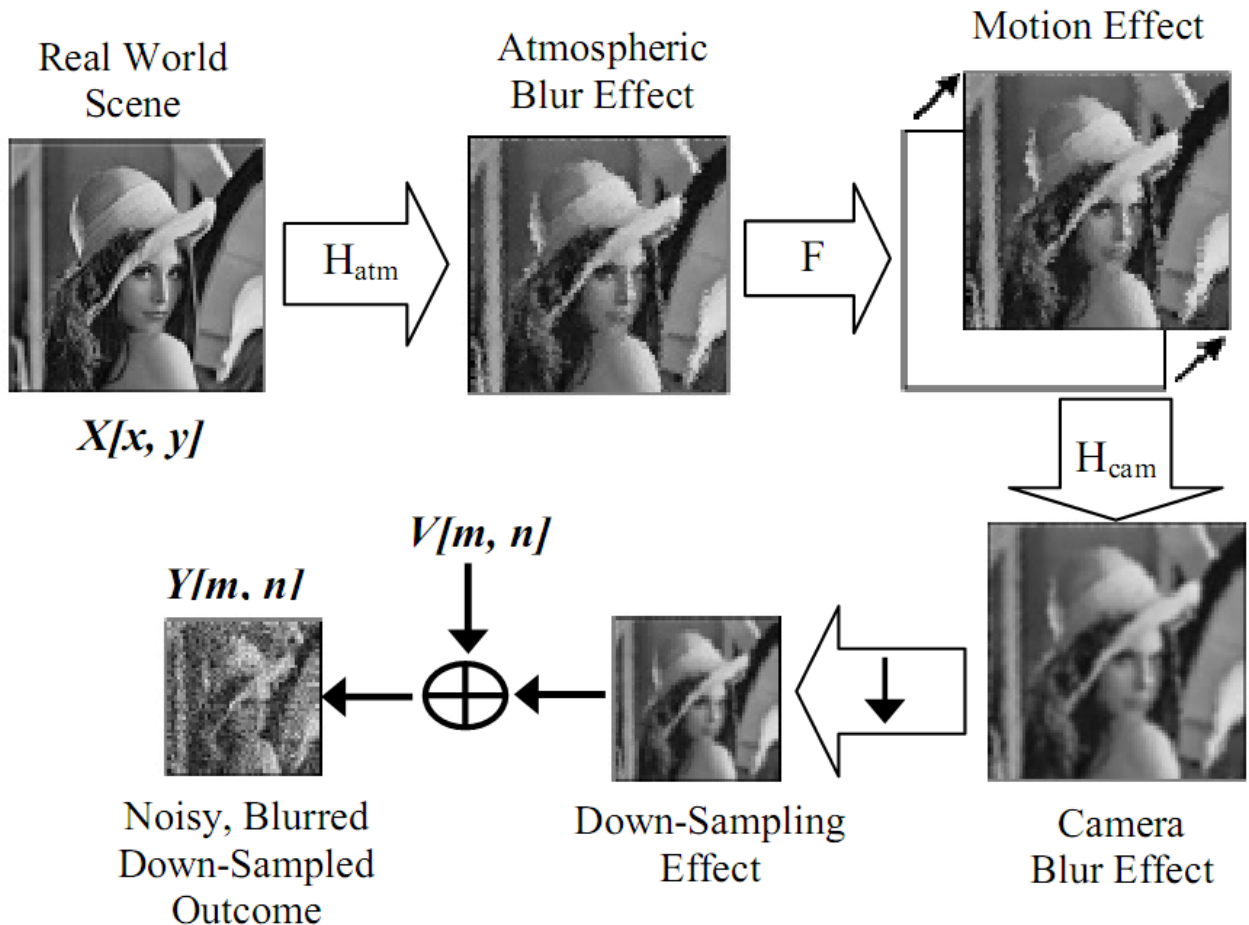

*Figure II.1 - Block diagram representation for four stages a,b,c,and d; where X(x,y) is the continuous intensity distribution of the scene, V [m, n] is the additive noise, and Y [m, n] is the resulting discrete low-quality image.*

a - Geometric Transformation (motion)

In order to have a unique super resolved image X, the coordinates system of X should be determined. A natural choice would be the coordinates system of one of the input images, enlarged by factor r, usually by two. The geometric transformation of X to the coordinates of the input images is computed by finding the motion between the input images. Actually, the motion estimation and model is explained in details in chapter 3.

b - Blur of an image

The main job of a digital camera is to transfer the scene from 3-dimentional world, where we live, to a 2-dimentional plane; and store it in a memory for us to be able to get use of or print out on demand.

This "projection" is not perfect. Blurring is a result of this imperfection. There are two kinds of blurring that occur to images. The images are blurred both by atmospheric turbulence and camera lens by continuous point spread functions (PSF) $H_{atm}(x, y)$ and $H_{cam}(x, y)$. The real image is convoluted with these spread functions to produce the blurred image.in other word, the digitized result of the camera blur is called "The PSF - Point Spread Function". In conventional image systems (such as video cameras), camera lens blur has more important effect than the atmospheric blur where we can add the two effect $H_{atm}$ and $H_{cam}$ to obtain one matrix called "H – the blur matrix ",where $H=H_{cam}H_{atm}$.

Image blur can usually be modeled by a convolution with some low-pass kernel. A kernel is $k \times k$ matrix which defines the effect of surrounding pixels of a target pixel. Along this work, we are going to consider 5 by 5 kernels.

ker1 = (1/19) * [0 0 1 0 0; 0 1 2 1 0; 1 2 3 2 1; 0 1 2 1 0; 0 0 1 0 0];
ker2 = (1/14) * [0 0 0 0 0; 0 1 2 1 0; 0 2 2 2 0; 0 1 2 1 0; 0 0 0 0 0];
ker3 = (1/16) * [0 0 0 0 0; 0 1 2 1 0; 0 2 4 2 0; 0 1 2 1 0; 0 0 0 0 0];
ker4 = (1/18) * [0 0 0 0 0; 0 1 2 1 0; 1 2 2 2 1; 0 1 2 1 0; 0 0 0 0 0];
ker5 = (1/25) * [1 1 1 1 1; 1 1 1 1 1; 1 1 1 1 1; 1 1 1 1 1; 1 1 1 1 1];…etc

Bellow we give examples of the blurring kernels that will be used in the following.

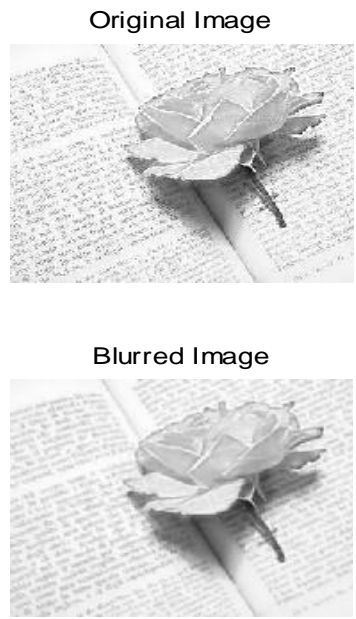

*Figure II.2 – convolute image original with ker2.*

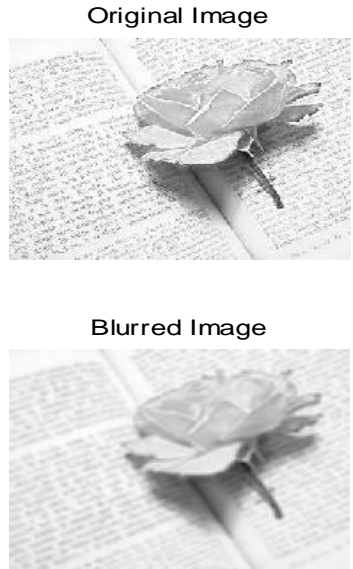

*Figure II.3 – convolute image original with ker5.*

c - The decimation

The decimation matrix is the matrix of the down sampling of the image, which changes from the HR image to a LR image. The only determinant of the decimation matrix is the decimation factor or the resolution factor r. The decimation operation is described in the figure below:

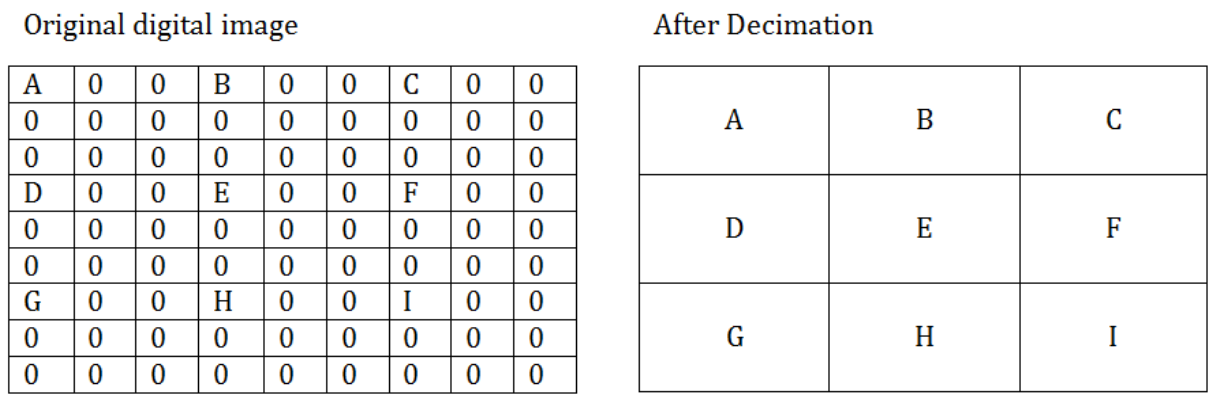

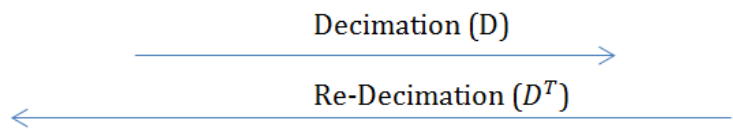

*Figure II.4 - The decimation operation.*

Each row in the D matrix contains all zeros except only one position having the value one. This position is the position of the pixel sampled. In figure II.1, we can see the effect of upsampling $D^T$ matrix on a 3by3 image and downsampling matrix D on the corresponding 9by9 upsampled image (resolution enhancement factor of three). In this figure, to give a better intuition, the image vectors are reshaped as matrices.

We note that $D^T$ copies the values from the LR grid to the HR grid after zero filling and D copies a selected set of pixels in HR grid back on the LR grid. Neither of these operators changes the pixel values.

d - Additive Noise

In SR, as in similar image processing tasks, it is usually assumed that the noise is additive, normally distributed with zero mean.

Under this assumption, the maximum likelihood solution is found by minimizing the error under Mahalanobis Norm (using estimated auto correlation matrix), or norm (assuming uncorrelated "white noise"). The minimum is found by using tools developed for large optimization problems under these norms, such as approximated kalman-filter, linear-equations solvers, etc. The assumption of normal distribution of the noise is not accurate in most of the cases, as most of the noise in the imaging process is non-gaussian (quantization, camera noise, etc.), but modeling it in a more realistic way would end in a very large and complex optimization problem which is usually hard to solve.

## II.2 – Modeling the imaging process

The major differences between most modern algorithms are in the optimization technique used for solving this set of equations, the constraints on X which are added to the system, and the modeling of the geometric transformation, blur and noise.

## II.2.1 - Schema Block of the Model

The first step in our study of SR image reconstruction problem is to formulate an observation model that relates the original HR image to the observed LR images.

Several observation models have been proposed in the literature, and they can be broadly divided into the models for still images and for video sequence. To present a basic concept of SR reconstruction techniques, we employ the observation model for still images, since it is rather straightforward to extend the still image model to the video sequence model.

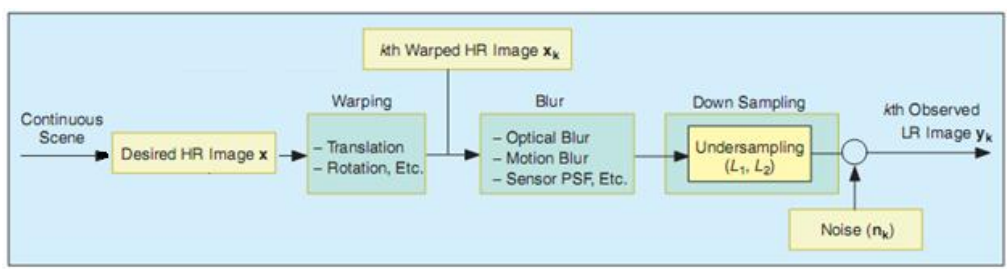

*Figure II.5 - Observation model relating LR images to HR images.*

## II.2.2 – Mathematical formulation

SR can be presented as a large sparse linear optimization problem, and solved using explicit iterative methods, in the presented framework a matrix-vector formulation is used in the analysis, but the implementation is by standard operations on images such as convolution, warping, sampling, etc.

Altering between the two formulations, a considerable speed up in the SR computation is achieved, by taking advantage of the two worlds: Implementing advanced gradient based optimization techniques (such as conjugate gradient), while computing the gradient in an efficient manner, using basic image operations, instead of sparse matrices multiplications.

In the analysis part, we consider the desired HR image of size $L_1 N_1 * L_2 N_2$ written in lexicographical notation as the vector $X =[x_1, x_2,....., x_N]^T$, where $N= L_1 N_1 \times L_2 N_2$. Namely, X is the ideal undegraded image that is sampled at or above the Nyquist rate from a continuous scene which is assumed to be band limited. Now, let the parameters $L_1$ and $L_2$ represent the down-sampling factors in the observation model for the horizontal and vertical directions, respectively. Thus, each observed LR image is of size $N_1 * N_2$. Let the $k^{th}$ LR image be denoted in lexicographic notation as $Y_k=[y_{k,1},y_{k,2},....,y_{k,M}]^T$ for k =1,2,...,p and $M=N_1*N_2$. Now, it is assumed that X remains constant during the acquisition of the multiple LR images, except for any motion and degradation allowed by the model. Therefore, the observed LR images result from warping, blurring, and subsampling operators performed on the HR image X.

Assuming that each LR image is corrupted by additive noise, we can then represent the observation model as [8], [9]

$$Y_k = D_k M_k B_k X + n_k \qquad (2)$$

Where :
1- $B_k$ represents a $L_1N_1L_2N_2 * L_1N_1L_2N_2$ blur matrix,
2- $M_k$ is a warp matrix (or motion matrix) of size $L_1N_1L_2N_2 * L_1N_1L_2N_2$,
3- $D_K$ is a $(N_1N_2)^2 * L_1N_1L_2N_2$ subsampling matrix,
4- and $n_k$ represents a lexicographically ordered noise vector.

As lightly different LR image acquisition model can be derived by discretizing a continuous warped, blurred scene [10]-[11]. In this case, the observation model must include the fractional pixels at the border of the blur support. Although there are some different considerations between this model and the one in (2), these models can be unified in a simple matrix-vector form since the LR pixels are defined as a weighted sum of the related HR pixels with additive noise [12]. Therefore, we can express these models without loss of generality as follows:

$$Y_k = W_k X + n_k, \text{ for } k=1, 2, ...., p \qquad (3)$$

where matrix $W_k$ of size $((N_1N_2)^2 * L_1N_1L_2N_2)$ represents, via blurring, motion, and subsampling, the contribution of HR pixels in X to the LR pixels in $Y_k$. Based on the observation model in (3), the aim of the SR image reconstruction is to estimate the HR image X from the LR images $Y_k$ for k=1, 2, ..., p

## II.2.3 - SR image Reconstruction Algorithms

The differences among the several proposed works are subject to what type of reconstruction method is employed, which observation model is assumed, in which particular domain (spatial or frequency) the algorithm is applied, what kind of methods is used to capture LR images, and so on. The technical report by Borman and Stevenson [13] provides a comprehensive and complete overview on the SR image reconstruction algorithms until around 1998, and a brief overview of the SR techniques appears in [14] and [15]. Based on the observation model in (3), existing SR algorithms are reviewed in the following sections.

There are many SR methods including non-uniform interpolation with fusion, frequency domain, deterministic and stochastic regularization, projection onto convex sets (POCS), and several other approaches, which are exactly solved by a huge number of papers.

In the following paragraph, we shall apply those tools and we add a regularize term to avoid the stair casing effects.

And now, we present an overview of the most popular algorithm.

## II.2.3.1 - Interpolation and Fusion

The first implemented algorithm to solve the SR issue is the interpolation and fusion method. Image interpolation has been used to increase the size of a single image. Single image interpolation cannot recover the high-frequency components lost or degraded during the LR sampling process. For this reason, image interpolation methods are not considered as SR techniques. To achieve further improvements in this field, the next step requires the utilization of multiple data sets in which additional data constraints from several observations of the same scene can be used. The fusion of information from various observations of the same scene allows us SR reconstruction of the scene [16].

This approach is the most intuitive method for SR image reconstruction. It passes through three stages presented in the figure below.

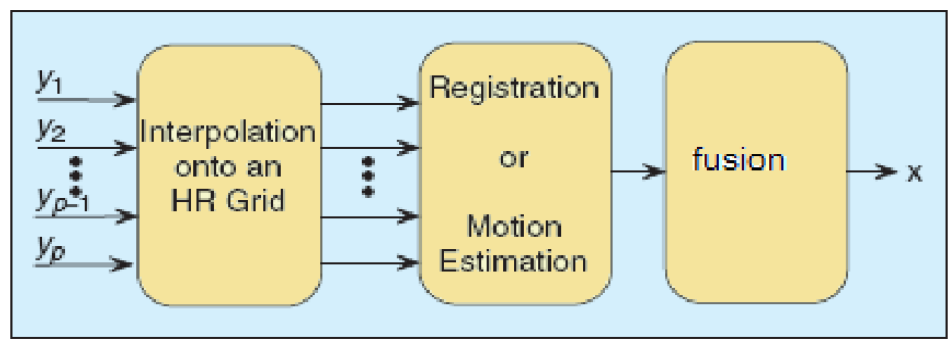

*Figure II.6 - Stages of algorithm.*

The first stage is to interpolate each of the LR images on a HR Grid. The interpolation method is done by first placing the LR pixel values on the HR grid with zero padding between the pixels where we show in figure II.2 in the second direction (re-decimation). Then we use an iterative technique to interpolate the 0-value pixels while keeping the initial values of the projected pixels unchanged over the whole iteration process. In a single iteration process, the target pixel is replaced by the sum of the values of the 2 neighboring pixels and its own value divided by three.

$$y(i,j)^{(n+1)} = \frac{1}{3} * (y(i-1,j)^{(n)} + y(i,j)^{(n)} + y(i+1,j)^{(n)}) \quad (4)$$

This is a single iteration of a pixel in a row direction. Where n is the iteration number and y is the interpolated HR image. Note that the iterations are only done on the pixels that where padded with zeros in the initial start of the interpolation. Moreover, the operation is done across the rows first and then across the columns. The convergence on this sequence is guaranteed.

The second stage is the estimation of relative motion; also known as registration. Some articles [17] had chosen to switch the first two steps as the registration comes first, but in our fusion model we can have sub-pixel displacement accuracy in estimating the translational motion between the frames. The sub pixel displacement is needed to give more accurate results for the registration. The correlation technique used in this part is the same described which MAD (mean average difference) cost function used to track the position of a block of a reference frame in the other frames. The output of this function would be a translational vector of the motion that the LR frames have with respect to the reference target frame.

The third step is the easiest but the most importance for the algorithm to gain success. We simply add the HR interpolated images and divide them by their number.

### II.2.3.2 - Projection on Convex Sets (POCS)

Low resolution images usually suffer from blurring caused by a sensor's PSF and additionally from aliasing caused by undersampling. Stark and Oskoui [10] have proposed a POCS technique that accounts for both the blurring introduced by the sensors as well as the effects of under sampling. In their model a low resolution image sequence is denoted by g ($m_1$, $m_2$, k). It is assumed that an estimate of the HR image at time k= $t_r$ is desired. A family of closed, convex constraint sets can be defined, one for each pixel within the low resolution image sequence

$$C_{tr} = (m_1, m_2, d) = \{y(n_1, n_2, t_r): | r^{(y)}(m_1, m_2, k) \leq \delta_0\} \quad (4)$$

Where
$$r^{(y)}(m_1, m_2, k) = g(m_1, m_2, k) - \sum_{(n1,n2)} y(n_1, n_2, t_r) h_{tr}(n_1, n_2, m_1, m_2, k) \quad (5)$$

is the residual associated with an arbitrary member, y, of the constraint set.

$h_{tr}$ combines the effect of the blur PSF and relative motion of object and sensor. The quantity $\delta_0$ is an a priori bound reflecting the statistical confidence with which the actual image, y, is a member of the set $C_{tr} = (m_1, m_2, k)$ This family of constraints is referred to as data consistency constraints. An estimate of the high- resolution version of the reference image is determined iteratively starting from some arbitrary initialization. Successive iterations are obtained by projecting the previous estimate onto the consistency set with an amplitude constraint set that restricts the gray levels of the estimate to the range [0, 255].

## II.3 - Performance metric

Images are subjective by nature; a good constructed image for some person is not good for another. A quantitative metric is an approach to obtain a comparison method. I have selected the mean average difference as the metric to evaluate the effectiveness of the system. The benefit of using this metric is that it provides a method for calculating the error that gives a numerical value that can be used to compare errors from different constructions regardless of the size of the image in question.

$$MAD = 1/L^2 \sum |X-\ddot{X}| \tag{6}$$

Where
$X$ – Original HR image
$\ddot{X}$ – Reconstructed HR image
L – Number of pixels along each axis (assumes square image)

## II.4 - Conclusion

A model of the SR was presented in this chapter. A HR image undergoes warping, blurring, down sampling, and finally noise addition and the LR images are the result. Remember that SR is the process on using several LR images to form one HR images. Therefore, implementation of the distortion blocks is a first step to solve the SR problem. Then we are going to take those algorithms and complete by a new adaptive terms (i.e regulariser) using MM algorithm, TV regulariser, and of course with motion.

# Chapter III

# Optical Flow

## III.1 – Introduction

With the preliminary work of Gibson in 1959 on "Motion parallax as a determinant of perceived depth" [18], where the terms optical change of the visual field", "optical motion" and "optical velocity" first appeared. Horn and Schunck, in 1981[19], define the "optical flow" as "the apparent motion of brightness patterns observed when a camera is moving relative to the objects being imaged". Their work has had a big influence on advanced image processing and motion estimation methods. Nowadays, the optical flow is used as a basic building block in many methods and applications in image and video processing, computer and machine vision and in related fields such as biomedical imaging, imaging for navigation etc.

For example, it is used in many video compression algorithms as a basic motion estimator. In biomedical imaging, it serves e.g. as building block for automatic computer tomographic image segmentation or for highly nonlinear image restoration for MRI images when the patient was moving during acquisition. In computer vision applications, such as mosaicking [20].

Numerous image registration techniques have been developed over past years [21]. Of these, optical flow [22] [23], and correlation-based methods are among the most popular. These methods are mainly developed to estimate the relative motion between a pair of frames. For cases where several images are to be registered with respect to each other (e.g. super-resolution applications), two simple strategies are commonly used. The first is to register all frames with respect to a single reference frame [24].

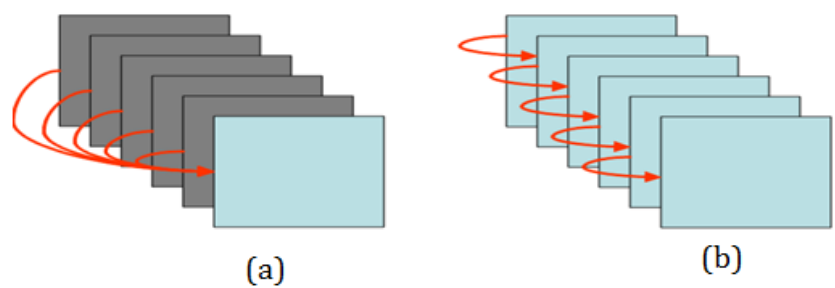

*Figure III.1 - Common strategies used for registering frames of a video sequence. (a) Fixed reference ("anchored") estimation. (b) Pairwise ("progressive") estimation.*

This may be called the anchoring approach, as illustrated in Figure III.1 (a). The choice of a reference or anchor frame is rather arbitrary, and can have a severe effect on the overall accuracy of the resulting estimates.

The other popular strategy is the progressive registration method [25], where images in the sequence are registered in pairs, with one image in each pair acting as the reference frame. For instance, taking a causal view with increasing index denoting time, the $i^{th}$ frame of the sequence is registered with respect to the $(i+1)^{th}$ frame and the $(i+1)^{th}$ frame is registered with respect to the $(i+2)^{th}$ frame, and so on, as illustrated in Figure III.1(b). The motion between an arbitrary pair of frames is computed from combined motion of the above incremental estimates. This method works best when the camera motion is smooth. However, in this method, the registration error between two "nearby" frames is accumulated and propagated when such values are used to compute motion between "far away" frames. Neither of the above approaches takes advantage of the important prior information available for the multi-frame motion estimation problem. This prior information constrains the estimated motion vector fields between any pair of frames to lay in a space whose geometry and structure, as we shall see in the next section, is conveniently described.

## III.2 - Methods for optical flow estimation

At present, we have a several groups of methods which solve the motion problem where each one has a special approach [24].
Bellow, there are some famous methods:

1. Lucas–Kanade Optical Flow Method – regarding image patches and an affine model for the flow field.
2. Horn–Schunck method – optimizing a functional based on residuals from the brightness constancy constraint, and a particular regularization term expressing the expected smoothness of the flow field    .
3. Buxton–Buxton method – based on a model of the motion of edges in image sequences
4. Black–Jepson method – coarse optical flow via correlation
5. General variational methods – a range of modifications/extensions of Horn–Schunck, using other data terms and other smoothness terms.

In this work, our interest is concentrated to the Horn-Schunck methods because the most frequently used methods pre-programmed in Matlab and Simulink are the Horn-Schunck and Lucas-Kanade ones. And between them, Horn-Schunck is more accurate.

## III.3 - Model for motion representation

The optical flow describes the direction and time rate of pixels in a time sequence of two consequent images. A two dimensional velocity vector, carrying information on the direction and the velocity of motion is assigned to each pixel in a given place of the picture. The optical flow could be used in situations when an observer (camera) is static and objects in the picture are moving. This is the case of railway crossing monitoring.
In order to detect the motion, one can use this classical model. If we denote by $I(x, y, t)$ the measured image intensity at position $(x, y, t)$ at time $t$, than $I'(x',y',t')$ measured image intensity at position $(x', y', t')$ at time $t'$ where $x'=x+\partial x$, $y'=y+\partial y$, $t'=t+\partial t$, represent in the following figure.

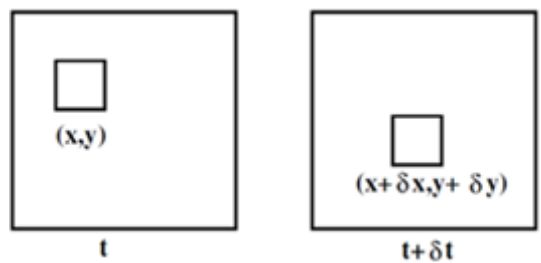

*Figure III.2 – Displacement of 2-D object.*

### III.3.1 -The optical flow constraint equation

A –The Brightness constraint.

Provided that the change of brightness intensity does not happen, in the neighborhood of the displaced pixel, we can use the following expressions [26].

$$I(x, y, t) = I'(x', y', t') = I(x+\partial x, y+\partial y, t+\partial t) \quad (1)$$

Thus, the optical flow constraint equation (OFCE) is obtained by using Taylor expansion on (1) and dropping its nonlinear terms, we obtain:

$$I(x+\partial x, y+\partial y, t+\partial t) = I(x, y, t) + \frac{\partial I}{\partial x}\partial x + \frac{\partial I}{\partial y}\partial y + \frac{\partial I}{\partial t}\partial t + H.O.T \quad (2)$$

From expressions (1) and (2) and with neglecting higher order terms (H.O.T.) we get

$$\frac{\partial I}{\partial x}\partial x + \frac{\partial I}{\partial y}\partial y + \frac{\partial I}{\partial t}\partial t = 0 \quad (3)$$

After modification $\quad \dfrac{\partial I}{\partial x}\dfrac{\partial x}{\partial t} + \dfrac{\partial I}{\partial y}\dfrac{\partial y}{\partial t} + \dfrac{\partial I}{\partial t}\dfrac{\partial t}{\partial t} = 0 \quad (4)$

or $\quad \dfrac{\partial I}{\partial x}v_x + \dfrac{\partial I}{\partial y}v_y + \dfrac{\partial I}{\partial t}1 = 0 \quad (5)$

Where $v_x$ and $v_y$ are components of pixel rate (optical flow) and $\partial I/\partial$ are derivations of the pixel brightness intensity in (x, y, t). Using common abbreviation of partial derivations the equation (5) after a slight modification holds

$$I_x.v_x + I_y.v_y = -I_t \quad (6)$$

or in a vector representation

$$(I_x, I_y).(v_x, v_y) = -I_t \quad (7)$$

or formally

$$\nabla I . \vec{v} = -I_t \quad (8)$$

Where $\nabla I$ is so-called the spatial gradient of brightness intensity and $\vec{v}$ is the optical flow (velocity vector) of image pixel and $I_t$ is the time derivation of the brightness intensity.
Thus the component of the movement in the direction of the brightness gradient $(I_x, I_y)$ equals

$$I_t / \sqrt{(I_x^2 + I_y^2)} \quad (9)$$

The formula (8) is keynote for optical flow calculation and is called 2-D Motion Constraint Equation or Gradient Constraint [26]. It represents one equation with two unknown quantities (the aperture problem).
Which means, we cannot determine optical flow uniquely only from such optical flow constraint equation.
Figure III.3 gives a geometrical explanation of the constraint equation.

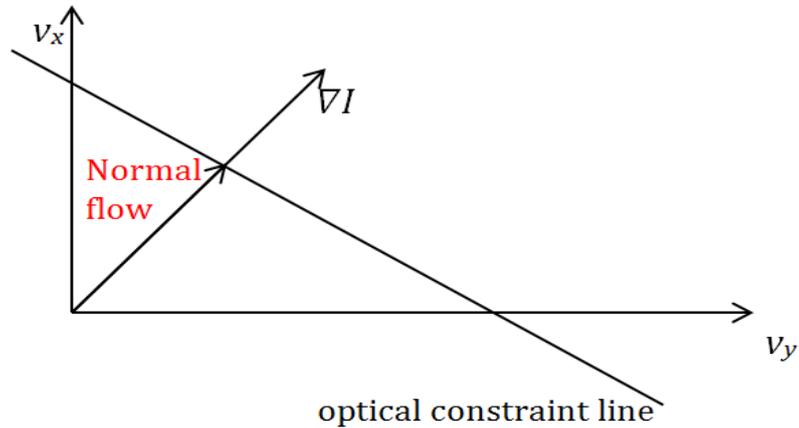

*Figure III.3 - Geometrical explanation of the optical flow constraint equation. The optical flow for a given image pixel can be any point on the constraint line in $v_x$, $v_y$ plane.*

B - The Smoothness constraint

If every point of the brightness pattern can move independently, there is little hope of recovering the velocities. More commonly we view opaque objects of finite size undergoing rigid motion or deformation. In this case neighboring points on the objects have similar velocities and the velocity field of the brightness patterns in the image varies smoothly almost everywhere. Discontinuities in flow can be expected where one object occludes another. An algorithm based on a smoothness constraint is likely to have difficulties with occluding edges as a result.
One way to express the additional constraint is to minimize the square of the magnitude of the gradient of the optical flow velocity:

$$(\partial v_x/\partial x)^2+(\partial v_x/\partial y)^2+(\partial v_y/\partial x)^2+(\partial v_y/\partial y)^2 \qquad (10)$$

Another measure of the smoothness of the optical flow field is the sum of the squares of the Laplacians of the x and y components of the flow. The Laplacians of $v_x$ and $v_y$ are defined as

$$\nabla^2 v_x = (\partial v_x/\partial x)^2+(\partial v_x/\partial y)^2 \quad and \quad \nabla^2 v_y = (\partial v_y/\partial x)^2+(\partial v_y/\partial y)^2 \qquad (11)$$

In simple situations, both Laplacians are zero. If the viewer translates parallel to a flat object, rotates about a line perpendicular to the surface or travels orthogonally to the surface, then the second partial derivatives of both $v_x$ and $v_y$ vanish (assuming perspective projection in the image formation).

## III.4 - Horn-Schunck model

This single equation (6) with two unknowns poses an aperture problem as describe by Tikhonov et al [27][28], some previously used terms such as quadratic smoother by Horn and Shunck [19], oriented smoother by Lukas Kanade[22] and anisotropic smoother by Brox [29]

The assumption made in this method is that optical flow varies smoothly, i.e., the variation of the optical flow field cannot be too big. Apparently, this is a global requirement for the whole image. Such smoothness constraint is indicated by the derivatives of optical flow, i.e., $\nabla v_x$ and $\nabla v_y$. The measure of departure from smoothness can be written by

$$e_s = \iint (\,||\nabla v_x||^2 + ||\nabla v_y||^2\,)dxdy \tag{12}$$

$$= \iint (\,(\partial v_x/\partial x)^2 + (\partial v_x/\partial y)^2 + (\partial v_y/\partial x)^2 + (\partial v_y/\partial y)^2\,)dxdy \tag{13}$$

The error of brightness constancy equation is:

$$e_c = \iint (\,I_x.v_x + I_y.v_y + I_t\,)^2 dxdy \tag{14}$$

So, we want to minimize:

$$e = e_c + \alpha\, e_s = \iint (\nabla I.v + I t)^2 + \alpha\,(||\nabla v_x||^2 + ||\nabla v_y||^2)dxdy \tag{15}$$

where domain of integrate is the region of the whole image, α expresses relative effect of the second added error term or scales the global smoothness term (typically α =1.0).

## III.4.1 - Approximation of partial derivatives and the Laplacian

We must estimate the derivatives of brightness from the discrete set of image brightness measurements available. It is important that the estimates of $I_x$, $I_y$ and $I_t$ be consistent. That is, they should all refer to the same point in the image at the same time. While there are many formulas for approximate differentiation , we will use a set which gives us an estimate of $I_x$, $I_y$, $I_t$, at a point in the center of a cube formed by eight measurements.

$$I_x \approx 1/4\{I_{i,j+1,t} - I_{i,j,t} + I_{i+1,j+1,t} - I_{i+1,j,t} + I_{i,j+1,t+1} - I_{i,j,t+1} + I_{i+1,j+1,t+1} - I_{i+1,j,t+1}\}$$
$$I_y \approx 1/4\{I_{i+1,j,t} - I_{i,j,k} + I_{i+1,j+1,t} - I_{i,j+1,k} + I_{i+1,j,t+1} - I_{i,j,t+1} + I_{i+1,j+1,t+1} - I_{i,j+1,t+1}\}$$
$$I_t \approx 1/4\{I_{i,j,t+1} - I_{i,j,t} + I_{i+1,j,t+1} - I_{i+1,j,t} + I_{i,j+1,t+1} - I_{i,j+1,t} + I_{i+1,j+1,t+1} - I_{i+1,j+1,t}\} \tag{16}$$

The three partial derivatives of images brightness at the center of the cube are each estimated from the average of first differences along four parallel edges of the cube.

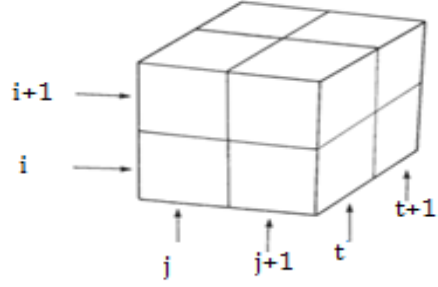

*Figure III.4 - Here the column index j corresponds to the x direction in the image, the row index i to the y direction. While t lies in the time direction.*

For the estimation of the Laplacean, we shall use

$$\nabla^2 v_x \approx k\,(\bar{v}_{x(i,j,t)} - v_{x(i,j,t)}) \quad \text{and} \quad \nabla^2 v_y \approx k\,(\bar{v}_{y(i,j,t)} - v_{y(i,j,t)}) \qquad (17)$$

Now, considering **k** = 3, and $\bar{v}_x$ and $\bar{v}_y$ will be the local median of $v_x$ and $v_y$, we shall approximate them by a 9-point stencil scheme, as follows

$$\begin{pmatrix} \dfrac{1}{12} & \dfrac{1}{6} & \dfrac{1}{12} \\ \dfrac{1}{6} & \dfrac{-1}{1} & \dfrac{1}{6} \\ \dfrac{1}{12} & \dfrac{1}{6} & \dfrac{1}{12} \end{pmatrix}$$

*Figure III.5 - The Laplacian is estimated by subtracting the value at a point from a weighted average of the values at neighboring points. Shown here are suitable weights by which values can be multiplied.*

Where the local averages $\bar{v}_x$ and $\bar{v}_y$ are defined as follows

$$\bar{v}_x(i,j,t) = \frac{1}{6}\left(v_{x(i-1,j,t)} + v_{x(i,j+1,t)}\right) + v_{x(i+1,j,t)} + v_{x(i,j-1,t)}) +$$
$$\frac{1}{12}\left(v_{x(i-1,j-1,t)} + v_{x(i-1,j+1,t)}\right) + v_{x(i+1,j+1,t)} + v_{x(i+1,j-1,t)})$$

$$\bar{v}_y(i,j,t) = \frac{1}{6}\left(v_{y(i-1,j,t)} + v_{y(i,j+1,t)}\right) + v_{y(i+1,j,t)} + v_{y(i,j-1,t)}) +$$
$$\frac{1}{12}\left(v_{y(i-1,j-1,t)} + v_{y(i-1,j+1,t)}\right) + v_{y(i+1,j+1,t)} + v_{y(i+1,j-1,t)}) \qquad (18)$$

## III.4.2 - Iterative solution

The problem then is to minimize the sum of the errors in the equation (15).

$$e = e_c + \alpha\, e_s = \iint (\nabla I \cdot v + I_t)^2 + \alpha\, (||\nabla v_x||^2 + ||\nabla v_y||^2) dx dy$$

So the minimization is to be accomplished by finding suitable values for the optical flow velocity $(v_x, v_y)$.

The relation (15) leads to the system of equations for whose solution it is convenient to use Jacobi or Gauss-Seidel iterative methods [30] but it would be very costly to solve these equations simultaneously by one of the standard methods. Horn-Schunck solve these two equations simultaneously by using block Gauss-Seidel relaxation in order to capture the coupling effect between them, expressed as [31]

$$v_x(i,j,t)^{<n+1>} = \bar{v}_x(i,j,t)^{<n>} - I_x \cdot (I_x \cdot \bar{v}_x + I_y \cdot \bar{v}_y + I_t) / (\alpha^2 + I_x^2 + I_y^2)$$
$$v_y(i,j,t)^{<n+1>} = \bar{v}_y(i,j,t)^{<n>} - I_y \cdot (I_x \cdot \bar{v}_x + I_y \cdot \bar{v}_y + I_t) / (\alpha^2 + I_x^2 + I_y^2) \quad (19)$$

where, $v_x(i,j,t)^{<n>}$, $v_y(i,j,t)^{<n>}$ is the velocity estimate for the pixel at (i,j) at time t, and $\bar{v}_x(i,j,t)^{<n>}$, $\bar{v}_y(i,j,t)^{<n>}$ is the neighborhood average of the optical velocity component $v_x$ and $v_y$. For k=0, the initial velocity is 0.

## III.5 - Conclusion

The advantages of the iterative Scheme include an ability to deal with more images per unit time and better estimates of optical flow velocities in certain regions. Areas in which the brightness gradient is small lead to uncertainly. This algorithm is implemented in the last chapter.

Chapter IV

Total variation-based image deconvolution:
a majorization-minimization approach.

## IV.1 – Introduction

SR refers to the reconstruction of a HR image from a set of blurred and noisy LR images which are sub-pixel shifted from each other.

Each LR image contains new information about the scene and SR aims at combining these to give a higher resolution image. SR image reconstruction algorithms investigate the relative sub-pixel motion information between multiple LR images and increase the spatial resolution by fusing them into a single frame.

The challenge in many linear inverse problems is that they are ill-posed, i.e., either the linear operator does not admit inverse or it is near singular, yielding highly noise sensitive solutions. To cope with the ill-posed nature of these problems, a large number of techniques have been developed, most of them under the regularization [32, 33] or the Bayesian frameworks [34].

The heart of the regularization and Bayesian approaches is the a priori knowledge expressed by the prior/regularization term.

## IV.2 - Total Variation (TV) Regularization-Based SR Model

TV regularization was introduced by Rudin, Osher, and Fatemi in [35] and has become popular in recent years [35, 36, 37, 38, 39, and 40]. Recently, the range of application of TV-based methods has been successfully extended to in painting, blind deconvolution, and processing of vector-valued images (e.g., color).

Arguably, the success of TV regularization relies on a good balance between the ability to describe piecewise smooth images and the complexity of the resulting algorithms.

In fact, the TV regularizer favors images of bounded variation, without penalizing possible discontinuities [41].this importance apparent in the follow figure

Most algorithms to compute TV-based estimates are fallen into two main categories [42]) solving the associated Euler-Lagrange equation, which is a non-linear partial differential equation (PDE) and 2) using methods based on duality.

The resulting algorithm for TV deblurring is related to iteratively reweighted least squares. Each iteration consists in minimizing a quadratic function, which is equivalent to solving a linear system. We note, however, that in the majorization-minimization (MM) framework we do not need to minimize the so-called majorizer function, but only to assure that it decreases.

Therefore, instead of computing the exact solution of a large system of equations, we simply run a few iterations of conjugate gradient (CG). For finite support convolutional kernels, the obtained algorithm has O(N) computational complexity. Experimental results illustrate the state-of-the-art competitiveness of the proposed approach.

In this work, we apply a MM method to design a new direct optimization technique formulated in the discrete domain. The MM rationale consists in replacing a difficult optimization problem by a sequence of simpler ones, usually by relying on convexity arguments. In this sense, MM is similar in spirit to expectation-maximization (EM). The advantage of the former resides in the flexibility in the design of the sequence of simpler optimization problems.

## IV.3 - MM Philosophy

The MM algorithm is an iterative algorithm that can be used to minimize or maximize a function. We focus on using it to maximize the log likelihood $l(Y; \theta)$ of observed data Y over model parameters $\theta$. Given a current guess $\theta^{(m)}$ for the parameters, the MM algorithm prescribes a minorizing function $h(\theta | \theta^{(m)})$ such that

$$h(\theta | \theta^{(m)}) \leq l(Y; \theta)$$
$$h(\theta^{(m)} | \theta^{(m)}) = l(Y; \theta^{(m)}) \qquad (1)$$

Figure IV.1 gives an example of log likelihood with accompanying minorizing function [43]. The minorizing function must be chosen such that the log likelihood dominates it everywhere except at $\theta^{(m)}$, where they are equal. If $h(\theta | \theta^{(m)})$ has derivatives everywhere, then the tangent $h'(\theta | \theta^{(m)}) = l'(Y; \theta)$ are also equal at $\theta = \theta^{(m)}$, and it is easy to see that choosing $\theta^{(m+1)}$ to maximize $h(\theta | \theta^{(m)})$ must also increase $l(Y; \theta)$. The challenge is to find a minorizing function $h(\theta | \theta^{(m)})$ that is easy to maximize.

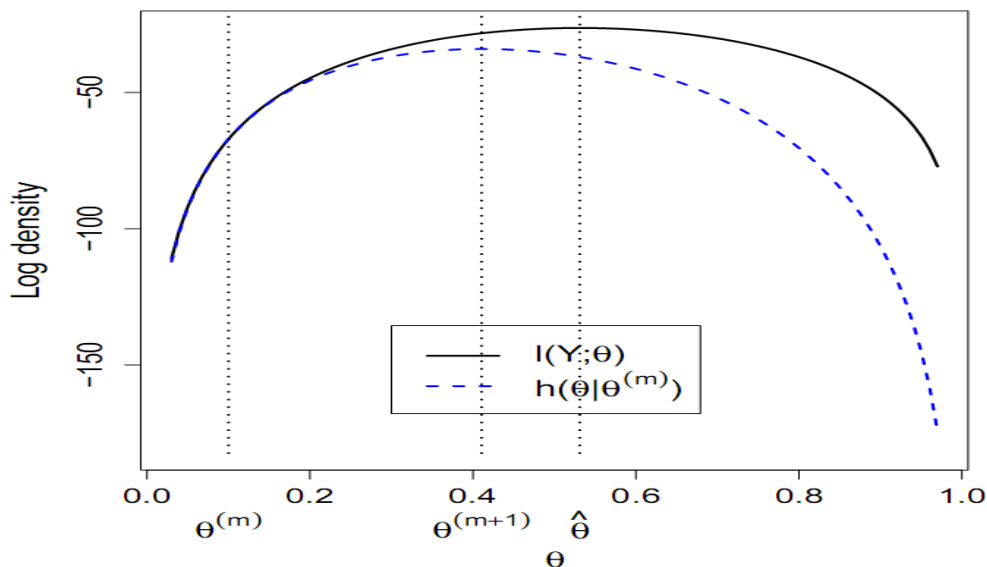

*Figure IV.1 - MM algorithm for minorization function $h(\theta | \theta^{(m)})$ of the observed log likelihood $l(Y; \theta)$.*

## IV.4 – Conjugate-Gradient (CG) method

To demonstrate the practical benefit of computing the gradient by image operations, a SR algorithm using the CG method was implemented. The CG method is an efficient method to solve linear systems defined by symmetric positive definite matrices.
Definition: Let Q be a symmetric and positive definite matrix. A vector set $\{V_i\}_{i=1}^n$ is Q-conjugate if

$$\vec{V}_i^T Q \vec{V}_j = 0, \quad \forall i \neq j \qquad (2)$$

A Q-conjugate set of vectors is linearly independent, and thus form a basis.
The solution of the linear equation is therefore a linear combination of these vectors. The coefficients $\alpha_i$ are very easily found:

$$Q\vec{X} = \vec{Y} \quad \Rightarrow \quad Q(\sum_{i=k}^{n} \alpha_k \vec{V}_k) = \vec{Y} \quad \Rightarrow \quad \alpha_k = \frac{\vec{V}_k^T \vec{Y}}{\vec{V}_k^T Q \vec{V}_k} \tag{3}$$

The CG algorithm iteratively creates a conjugate basis, by converting the gradients computed each iteration to vectors which are Q-conjugate to the previous ones (e.g. Graham-Shmidt procedure). Its convergence to the solution in steps is guaranteed, but this is irrelevant to the SR problem, since, the matrix size in SR, is huge. Still, the convergence rate of the CG is superior to steepest descent methods. Below is an implementation of CG.

CG(X, Y, Q, ε, mMax) solves QX = Y, ε, mMax limit the number of iterations
1. r = Y-QX, $\rho_0$ = || r ||², m=1
2. Do While $\sqrt{\rho_0 - 1}$ > ε|| Y ||² and m ≤ kmax
    a) If m=1 then p=r
       else β = $\frac{\rho_{m-1}}{\rho_{m-2}}$ and p = r + βp
    b) W=Qp
    c) α = $\frac{\rho_{m-1}}{p^T w}$
    d) X = X + αp , r = r − αw, $\rho_m$ = || r ||², m= m +1

In the CG implementation to SR, the input includes the LR images, ε, mMax and the estimated blur function.

First the motion between the input images is computed, and an initial estimate to the solution $X_H$ {0} is set, for example the average of the bilinearly upsampled and aligned input images. For more details please read [20].

## IV.5 - Problem Formulation

In this section, we divide the theory in two parts: the first, take the model for one observation image .the second one, we take the SR model for continuous framework.

## IV.5.1 – TV based single image deconvolution

From chapter 3, we consider the linear observation model:

$$Y = D\ B\ MX + n \tag{4}$$

In other term, we can writhe:

$$Y = HX + n \tag{5}$$

where X and Y denote vectors containing the true and the observed image gray levels, respectively, arranged in column lexicographic ordering, H is the observation matrix and n is a sample of a zero-mean white Gaussian noise vector with covariance $\alpha^2$ I (where I denotes the identity matrix).

As in many recent publications [35, 36, 37, 38, 39, 40], we adopt the TV regularizer to handle the ill-posed nature of the problem of inferring X. This amounts to computing the herein termed TV estimate, which is given by

$$\hat{X} = argmin_x L(X) \tag{6}$$

with

$$L(X) = ||Y-HX||^2 + \lambda TV(X) \tag{7}$$

where λ controls the relative weights of the data compliance and regularization terms. Since we are assuming, from the beginning, that images are defined on discrete domains.
The objective function L(X) is convex, although not strictly so. Nevertheless, its minimization represents a significant numerical optimization challenge, owing to the non-differentiability of TV(X).

A big number of researches are taken the TV regularizer and we can classify it in forms:
1- The classic term:

$$TV(X) = \sum_i \sqrt{(\Delta_i^h X)^2 + (\Delta_i^v X)^2} \tag{8}$$

Where $\Delta_i^h$ and $\Delta_i^v$ are linear operators corresponding to horizontal and vertical first order differences, at pixel i, respectively; i.e., $\Delta_i^h X \equiv x_i - x_{j_i}$ (where $j_i$ is the first order neighbor to the left of i) and $\Delta_i^v X \equiv x_i - x_{k_i}$ (where $k_i$ is the first order neighbor above i).
The function behaves as the classical total variation near edges and in flat regions it induces an isotropic Laplacian diffusion. We can give the well-known example of minimal surfaces regularizing term to obtain an update form.
2- The update term propose by [44]:

$$TV(X) = \sum_i \sqrt{((\Delta_i^h X)^2 + (\Delta_i^v X)^2 + 1)} - 1 \tag{9}$$

Another idea proposed in [45, 46] is to find a function

$$TV(X) = \sum_i \sqrt{(\Delta_i^h X)^2 + (\Delta_i^v X)^2}^{p(\sqrt{(\Delta_i^h X)^2 + (\Delta_i^v X)^2})} \tag{10}$$

Where

$$p(s) = \frac{2 + \log(1+s)}{1 + \log(1+s)} \tag{11}$$

In the next section, we introduce a new optimization algorithm, fully developed on the discrete domain, which is simple and yet computationally efficient.

## IV.5.1.1 - An MM approach to TV deconvolution

Let $X^{(t)}$ denote the current image iterate and $Q(X|X^{(t)})$ a function that satisfies the following two conditions:

$$L(X^{(t)}) = Q(X^{(t)}|X^{(t)}) \tag{12}$$
$$L(X) \leq Q(X|X^{(t)}), X \neq X^{(t)}, \tag{13}$$

i.e, $Q(X|X^{(t)})$, as a function of X, majorizes (i.e., upper bounds) L(X). Suppose now that $X^{(t+1)}$ is obtained by

$$X^{(t+1)} = argmin_x Q(X|X^{(t)}) \tag{14}$$

then

$$L(X^{(t+1)}) \leq Q(X^{(t+1)}|x(t)) \leq Q(X^{(t)}|X^{(t)}) = L(X^{(t)}) \tag{15}$$

where the left hand inequality follows from the definition of Q and the right hand inequality from the definition of $X^{(t+1)}$.

The sequence $L(X^{(t)})$, for t = 1, 2, ... ,is, therefore, non-increasing. Under mild conditions, namely assuming that Q(X|X') is continuous in both X and X', all limit points of the MM sequence $L(X^{(t)})$ are stationary points of L, and $L(X^{(t)})$ converges monotonically to L* =L(X*), for some stationary point X*. If, in addition, L is strictly convex, then $X^{(t)}$ converges to the global minimum of L. The proof of these properties parallels that of the EM algorithm, which can be found in [47].

Observe that in order to have $L(X^{(t+1)}) \leq L(X^{(t)})$, it is not necessary to minimize $Q(X|X^{(t)})$ w.r.t x, but only to assure that $Q(X^{(t+1)}|X^{(t)}) \leq Q(X^{(t)}|X^{(t)})$. This property has a relevant impact, namely when the minimum of Q cannot be found exactly or it is hard to compute.

The majorization relation between functions is closed under sums, products by nonnegative constants, limits, and composition with increasing functions [49]. These properties allow us to tailor good bound functions Q, a crucial step in designing MM algorithms. This topic is extensively addressed in [19]

## IV.5.1.2 - A quadratic bound function for L(X)

We now derive a quadratic bound function for L(X). The motivation is twofold: first, minimizing quadratic functions is equivalent to solving linear systems; second, we do not need to solve exactly each linear system, but simply to decrease the associated quadratic function, which can be achieved by running a few steps of the CG algorithm.

Note that the term $||Y-HX||^2$, present in the definition of L in (33), is already quadratic. Let us then focus our attention on each term of TV(X) given by classic form. Using the fact that

$$\sqrt{x} \leq \sqrt{x_0} + \frac{1}{2\sqrt{x_0}}(x - x_0) \tag{16}$$

for any $x \geq 0$ and $x_0 > 0$, it follows that the function $Q_{TV}$ defined as

$$Q_{TV}(X|X^{(t)}) = TV(X^{(t)}) + \frac{\lambda}{2}\sum_i \frac{[(\Delta_i^h X)^2 - (\Delta_i^h X^{(t)})^2]}{\sqrt{(\Delta_i^h X^{(t)})^2 + (\Delta_i^v X^{(t)})^2}}$$
$$+ \frac{\lambda}{2}\sum_i \frac{[(\Delta_i^v X)^2 - (\Delta_i^v X^{(t)})^2]}{\sqrt{(\Delta_i^h X^{(t)})^2 + (\Delta_i^v X^{(t)})^2}} \quad (17)$$

satisfies $TV(X) \leq Q_{TV}(X|X^{(t)})$, for $x \neq x(t)$, and $TV(x) = Q_{TV}(x|x^{(t)})$, for $x=x(t)$. Function $Q_{TV}(x|x^{(t)})$ is thus a quadratic majorizer for $TV(x)$.

Let $\mathbf{D}^h$ and $\mathbf{D}^v$ denote matrices such that $\mathbf{D}^h X$ and $\mathbf{D}^v X$ yield the first order horizontal and vertical differences, respectively.

Define also $\mathbf{W}^{(t)} = \text{diag}(w^{(t)}; w^{(t)})$, where

$$W^{(t)} = \frac{\frac{\lambda}{2}}{\sqrt{(\Delta_i^h X^{(t)})^2 + (\Delta_i^v X^{(t)})^2}} \quad i = 1, 2, \ldots \quad (18)$$

With these definitions, $Q_{TV}(X|X^{(t)})$ can be written in a compact notation as

$$Q_{TV}(X|X^{(t)}) = X^T D^T W^{(t)} D X + c^{te} \quad (19)$$

where $\mathbf{D} = [(\mathbf{D}^h)^T \ (\mathbf{D}^v)^T]^T$, and $c^{te}$ stands for a constant.

Given that the first term of $L(x)$ in (3) is quadratic, a quadratic bound function for L is thus

$$Q(X|X^{(t)}) = ||Y - HX||^2 + Q_{TV}(X|X^{(t)}) \quad (20)$$

We stress that matrix $W^{(t)}$, present in $Q_{TV}(X|X^{(t)})$, is computed from $X^{(t)}$. The minimization of (20) leads to the following update equation:

$$X^{(t+1)} = (H^T H + D^T W^{(t)} D)^{-1} H^T Y \quad (21)$$

Obtaining $X^{(t+1)}$ via (21) is hard from the computational point of view, as it amounts to solving the huge linear system $A^{(t)}X = Y'$, where $A = H^T H + D^T W^{(t)} D$ and $Y' = H^T Y$. We tackle this difficulty by replacing the minimization of $Q(X|X^{(t)})$ with a few CG iterations, thus assuring the decrease of $Q(X|X^{(t)})$, with respect to X. The resulting scheme is still an MM algorithm.

**Algorithm 1:**
**MM Algorithm for TV deconvolution [49]**

Initialization: $X_0 = Y'$
1: for t := 0 to StopRule do
2: $W^{(t)} := \text{diag}[w^{(t)} \ w^{(t)}]$ {$w^{(t)}$ given by (18)}
3: $X^{(t+1)} := X^{(t)}$
4: while $||A^{(t)} X^{(t+1)} - Y'|| \geq \varepsilon ||Y'||$ do
5: $X^{(t+1)} :=$ next CG iteration
6: end while
7: end for

Algorithm 1 shows the pseudo-code for the proposed MM scheme. Line 2 implements the majorization step; lines 3 to 6 decrease $Q(X|X^{(t)})$. The parameter ε in line 4 implicitly controls the number of CG iterations.

## IV.5.2 – TV approach for SP

In this section we do the same work in section IV.5.1 but with sequence of observation images respecting the following model:

$$Y_k = D_k M_k B_k X + n_k \qquad (22)$$

Where :
5- $B_k$ represents a $L_1N_1L_2N_2 * L_1N_1L_2N_2$ blur matrix,
6- $M_k$ is a warp matrix (or motion matrix) of size $L_1N_1L_2N_2 * L_1N_1L_2N_2$,
7- $D_K$ is a $(N_1N_2)^2 * L_1N_1L_2N_2$ subsampling matrix,
8- and $n_k$ represents a lexicographically ordered noise vector.

The discrete energy to be minimized is given by λ

$$L(X) = ||H_k X - Y_k||^2 + \lambda TV(X) \quad \text{for k=1, 2, ..., N} \qquad (23)$$

where TV(X) is given by classic term or update one, ||.|| denotes the Euclidean norm , k the number of LR image, and λ is the Lagrangian $c^{te}$ where k = 1,..., N
If we follow the same scheme of section IV.5.1, we obtain the iteration solution of HR image X.

$$X^{(t+1)} = \left(\sum_{k=1}^{N}(H_k^T H_k) + D^T W^{(t)} D\right)^{-1} \sum_{k=1}^{N}(H_k^T Y_k) \qquad (24)$$

where

$$H_k = D_k M_k B_k \quad \text{for k=1,2, ..., N} \qquad (25)$$

Obtaining $X^{(t+1)}$ via (24) is hard from the computational point of view, as it amounts to solving the huge linear system $A^{(t)} X = Y'$, where $A = \left(\sum_{k=1}^{N}(H_k^T H_k) + D^T W^{(t)} D\right)$ and $Y' = \sum_{k=1}^{N}(H_k^T Y_k)$. We tackle this difficulty by replacing the minimization of $Q(X_k|X_k^{(t)})$ with a few CG iterations, thus assuring the decrease of $Q(X_k|X_k^{(t)})$ with respect to x. The resulting scheme is still similar to algorithm 1 in previous part.

## IV.6 - Conclusion
We have developed a new MM algorithm for image and video deconvolution under total variation regularization with tow classic and update term. The complexity of the algorithm is O(k*N) for finite support convolution kernels, where N is the number of image pixels and k is number of image LR. The efficient of MM approach is clear; this algorithm transfer a difficult problem to iterative sample one. And the efficient of TV regulazer is tested in last chapter.

# Chapter V
# Implementation using MATLAB

## V.1 – Introduction

MATLAB powerful tools for computations with vectors and matrices make this package well suited for solving typical problems of linear programming.
From this reason, we chose the MATLAB program to compute our algorithm.
In the following test, we use the computer specification: Sony VAIO Quad core intel® Core™ i7 processor (1.6 GHz) with turbo Boost up to (2.8 GHz), 4GB of RAM.
This chapter divides in tow big part:
1- Experiment the algorithm Horn Shunk to estimate motion.
2-Experiment the concepts of SR image reconstruct where is treatment in chapter IV.

## V.2 - Horn-Schunck estimation

We have implemented Horn-Schunck algorithms using their best appearance adjustments. The testing sequences are obtained from a moving box with size 240x320*3 on a conveyer belt with various speeds of vertical displacement.
The estimated optical flow use **HornSunck1.m** function (appendix A); results are illustrated in Figure V.1

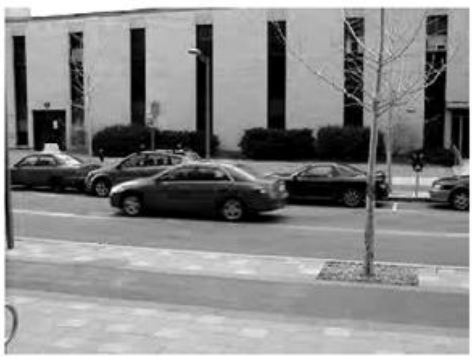
refrence image frame1

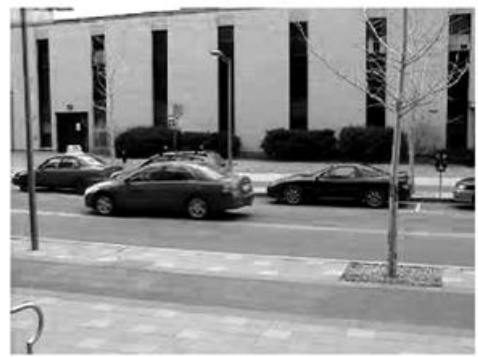
refrence image frame2

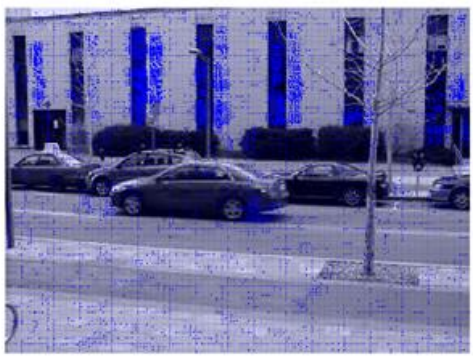
Optical flow: respect HS algorithm with alpha=1, and iteration=20

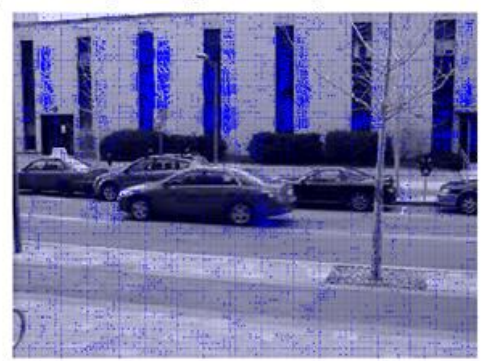
Optical flow: respect HS algorithm with alpha=1, and iteration=50

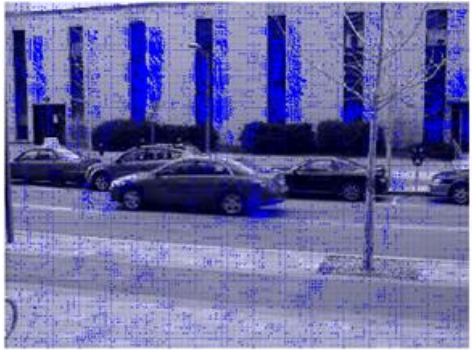

*Figure V.1 - Here we present two images from video and the result of Horn Schunk algorithm respectively with iteration equal 20, 50, and 100.*

Because our interested of the values of vectors velocity; we apply that algorithm of Horn Schunk in virtual image case and we decide the shifting values to compare with the estimated one.
Original image has size 128x128, rectangle with object and black background

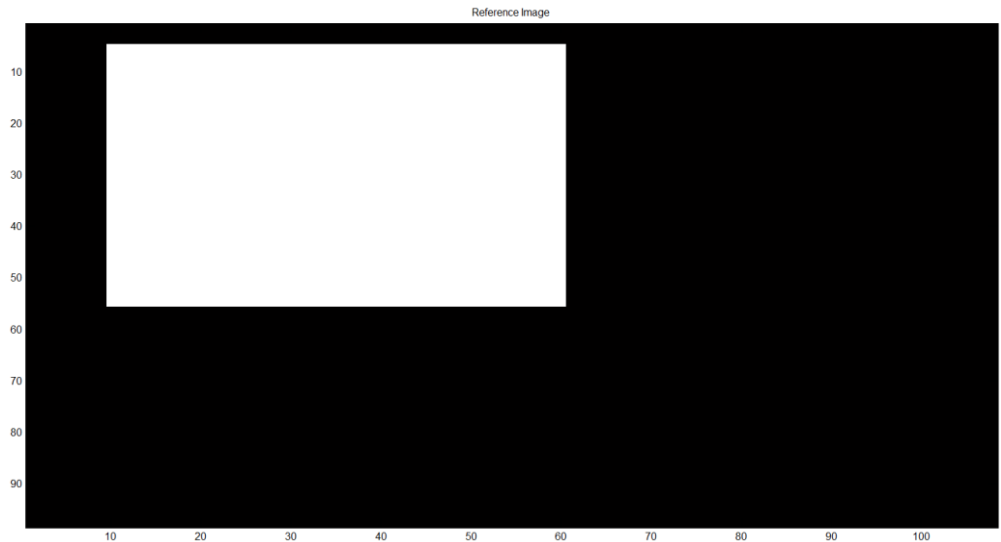

*Figure V.2 - The virtual image create by MATLAB, has size 128x128.*

We do a shifting in tow direction i.e in vertical and horizontal where horizontal shift = 10 (along columns) and vertical shift = 15 (along line).
In MATLAB command windows, we are:
True shifts:

ans = 10    15

Estimated shifts:

ans = 10.0    15.0

And the shifted image is:

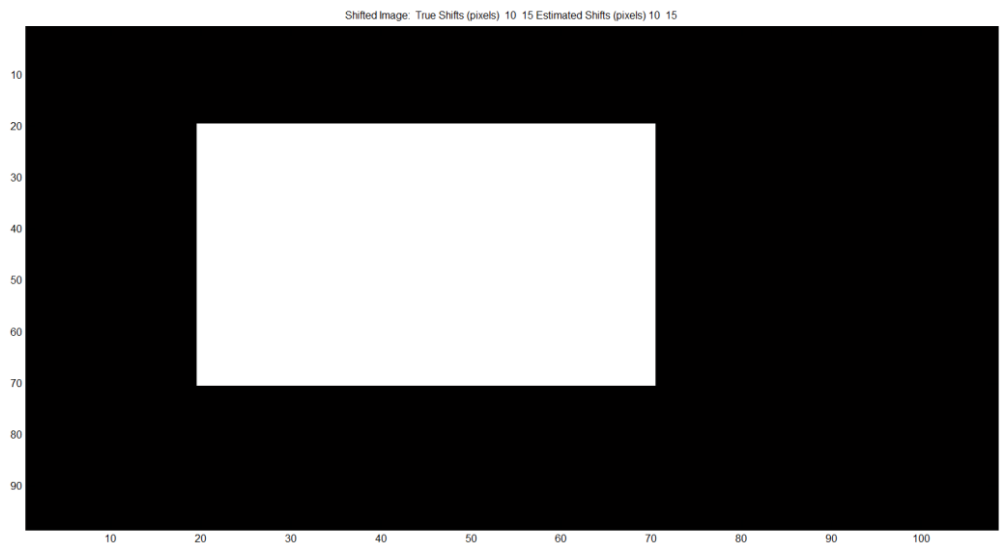

*Figure V.3 - The virtual shifted image.*

Now return to a real case, for we take image in figure V.1 where the true shift between them is horizontal shift = 12 and vertical shift = 1.
In MATLAB command windows, we are:

True shifts:

ans = 12    1

Estimated shifts:

ans = 11.6097    0.7759
And the shifted image is:

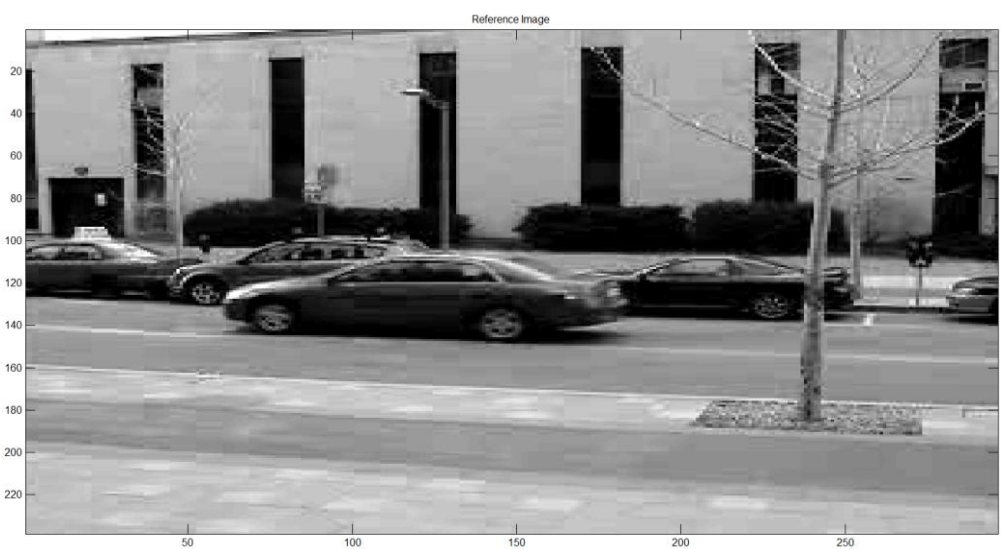

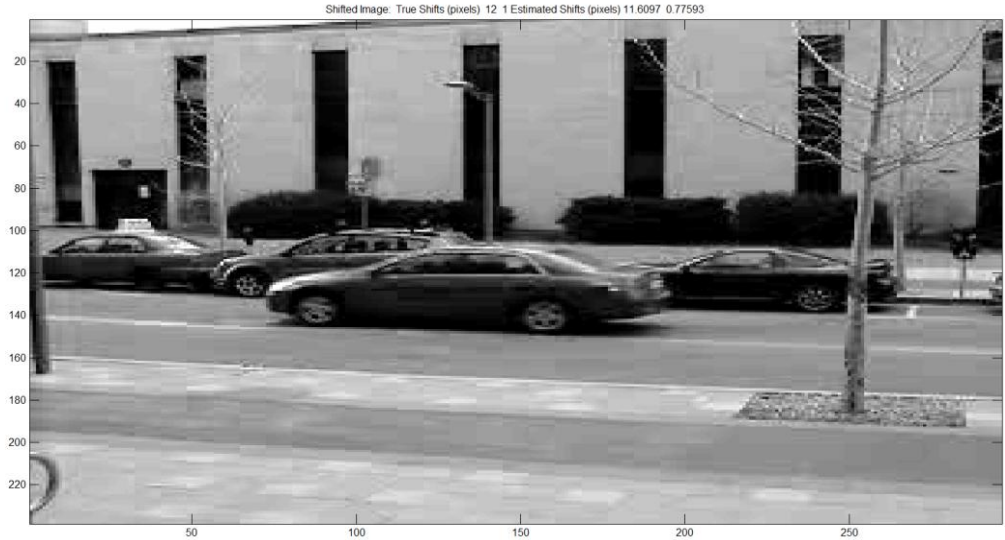

*Figure V.4 – The first image is reference and the second one is result of shift 12, 1 respectively horizontal and vertical.*

## V.3 - Implementation of SR model

In this chapter, we are going to implement each of the above matrices to create blurred, noisy, warping and decimated LR images from a single HR image.

### V.3.1 - The Blur matrix B, Motion matrix M, and Decimation matrix D

As defined in a previous chapter, blurring is defined by a convolution kernel (PSF) which would be convolved with the matrix image. Since convolution cannot be represented by an ordinary matrix multiplication operator, we placed an image in lexicography order. From the equation above,

$$Y_k = D_k M_k B_k X + n_k \quad k=1, 2, …, N \tag{1}$$

We will make an assumption that all the N-LR images undergo the blurring (which is true in case we are using the same camera). Therefore, $B_1, B_2, B_3, B_4, …, B_N$

Since B is applied to the HR image vector ($X$), then the dimensions of B is [$L_1N_1L_2N_2$* $L_1N_1L_2N_2$]. For example, when we are using an image 256 by 256 pixels, the size of blur matrix B would be 65536 by 65536. This demonstrates the need of powerful machines of multiprocessors in dealing with image processing problems. All our work had been implemented using an ordinary personal computer of limited capabilities. But this can be extended to include higher resolution pictures.

Each row in the matrix B can be programmed to give any combination of the pixels in the whole image. Correctly filling the rows of B, leads to any desired function of linear combination of the image pixel values. To create the blurring matrix B, the needed inputs are the blur kernel (or PSF of dimensions 5 by 5), and the high resolution frame size.

Motion matrix M is applied to the HR image vector (B*X), then the dimensions of M is [$L_1N_1L_2N_2$ * $L_1N_1L_2N_2$].

D is a $(N_1N_2)^2$ * $L_1N_1L_2N_2$ subsampling matrix.

The noise vector $n$ will be considered as Gaussian noise vector having zero mean. Moreover, an image noise can be also estimated using the signal to noise ratio (SNR). The equation to calculate the SNR is as follows:

$$\text{SNR} = 10\log_{10}\frac{\sigma^2}{\sigma_N^2} = 10\log_{10}\frac{\frac{1}{L^2}\sum_1^{LxL}(Y-\mu Y)^2}{\sigma_N^2} \quad (2)$$

Where

$\sigma$: Standard deviation of the clean image Y without noise.

$\mu_Y$: Average of the image.

$\sigma_N$: Standard deviation of the noise.

The unit of the SNR metric is the decibel (dB). In the above experiment, the Gaussian noise having $\sigma = 1$ given by use randn() function in MATLAB.

We are now able to test our forward model.

The experiments are done on several pictures shown below. From one high resolution image, we have made 3 blurred, decimated, and noisy images.

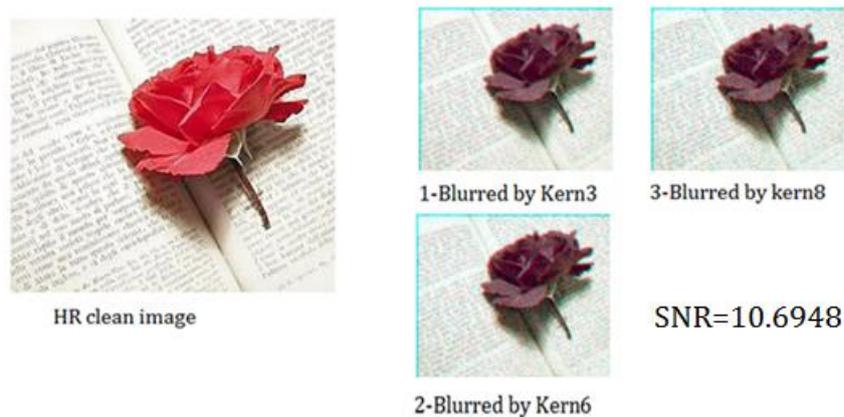

*Figure V.5 – Forward model. Blurred, Decimate, Motionless and Noisy images.*

Other example with motion

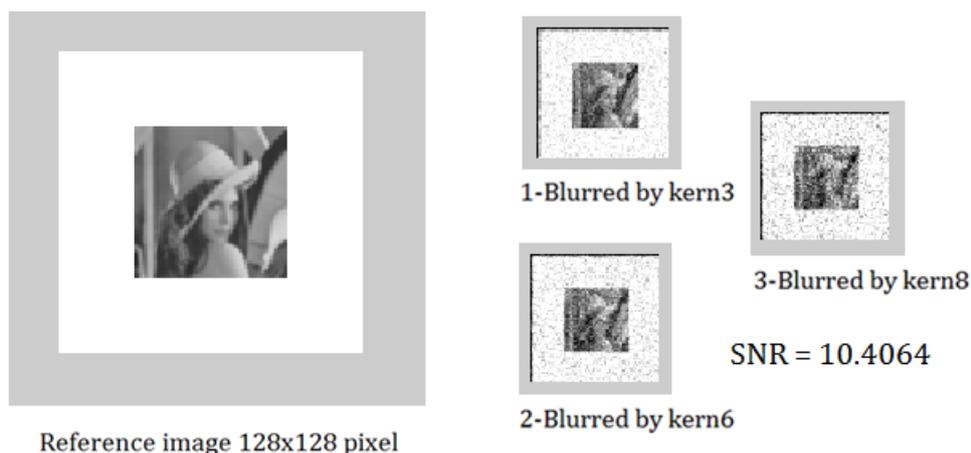

*Figure V.6 – Forward model. Blurred, Decimate, Motion and Noisy images.*

This two examples represents the forward model from HR to serial LR image, but actually a have a problem with of motion matrix when I use image size 128x128 pixels where the size of matrix M equal 16384x16384 is a huge matrix and it is impossible to save it with format (.mat). For this reason we reduce the size of HR from 128x128 to 64x64 and the M matrix has 4096x4096.

## V.4 - Experiments TV regularization via MM algorithm

We have presented above the minimization algorithm and the corresponding method to solve the nonlinear optimization problem (view chapter VI: **formula 23**). In other term minimize the error function to obtain HR image X.

$$L(X) = || H_k X - Y_k ||^2 + \lambda TV(X) \quad \text{for k=1, 2, ..., N} \quad (3)$$

Our final step would be to test how effective is our algorithm (i.e chapter VI: **Algorithm 1 MM Algorithm for TV deconvolution** ) in implementation.
To implement this algorithm, we have two functions doing the SR reconstruct:
1-Deconvsuper.m
2- Deconvsuperplusone.m
Those functions are uses four LR blurred, decimated and noisy images to create one SR image.
The Inputs of this function are:
8 LR images obtain by previous section where the blur effect , motion and decimation are modeling obtain by the output of function **Imagetransformation** (inputs of this function are: decimation factor+kernel+Motion matrix). The candidate Blur Kernels are: ker1,…,ker8 declared in bigmain.

The output of imagetransformation.m is a matrix H where H*X give the LR image. At the LR image result, we add a Gaussian noise having $\sigma = 1$ given by use randn() function in MATLAB.

We have considered two images and multispectral image of size 64x64 as test images. The HR input images are shifted, rotated, blurred and additive noise. The result is $Y_k$(k=1, …, 8), The images are deconvoluted by using deconsuper or deconsuperplusone. In fact, the function deconsuper.m use the classic form of TV regulazer and deconsuperplusone.m use the update form of TV regulazer (view section IV.5.1 in chapter VI).

1- First image lena256.tif with motionless and without regulazer term.

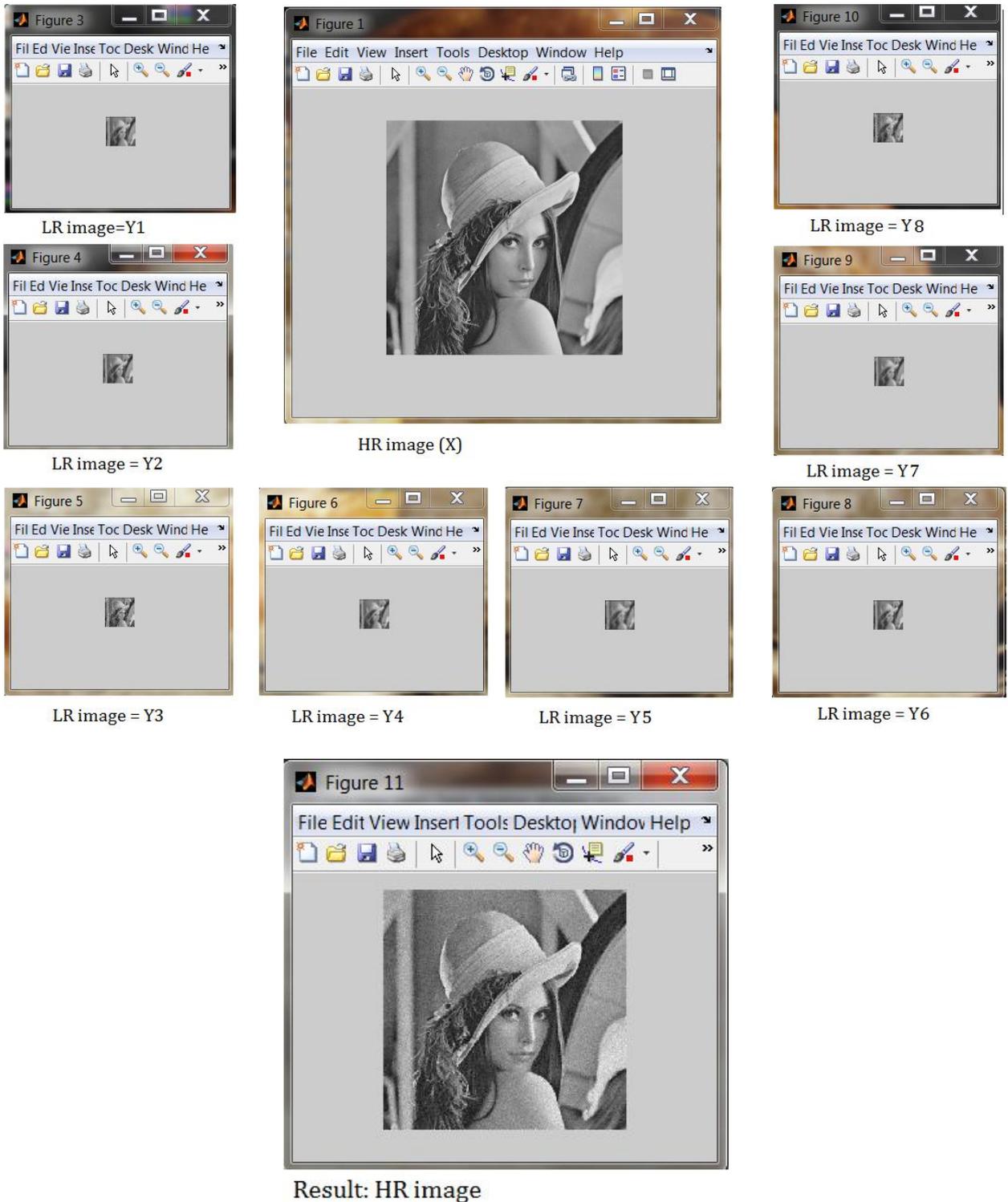

*Figure V.7 - HR image is the reference image with size 256x256, other images are the LR images respectively Y1,Y2,...,Y8 with size 64x64. the result is HR image with the same size of reference image.*

2- Second image lena128.tif with white edge. Motion and with regulazer are existed here.

The eight LR obtains have everyone a specific variance where the variances are respectively represented:

Varnoise1 = 5.8554e-004; Varnoise2 = 0.1872 ; Varnoise3 =    18.7195; Varnoise4 = 5.8942; Varnoise5 = 10.3357; Varnoise6 = 1.8760 ; Varnoise7 =  0.1840 ; Varnoise8 = 18.4117

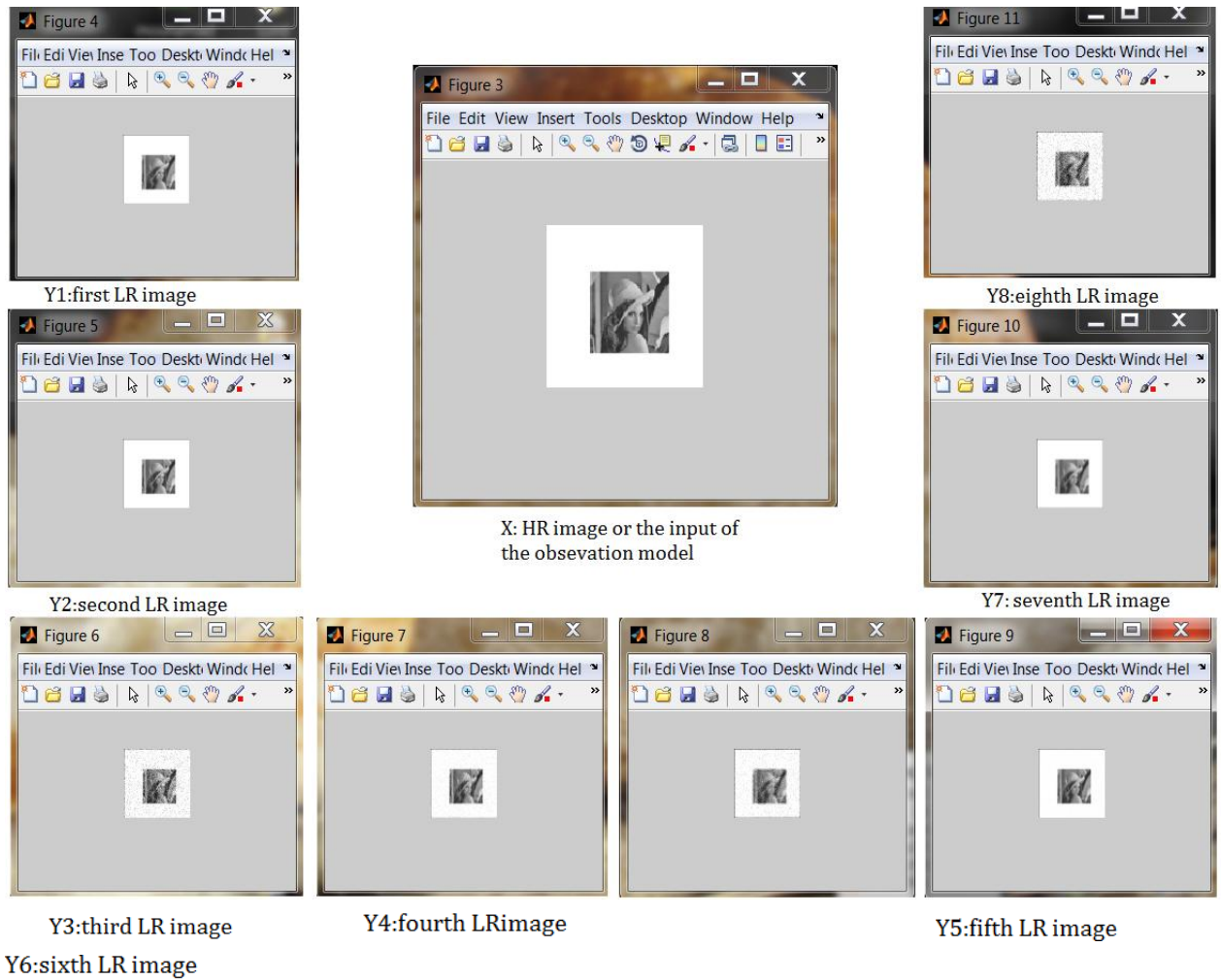

*Figure V.8 – X: HR image is the reference image with size 128x128, other images are the LR images respectively Y1, Y2, …, Y8 with size 64x64.*

The eight LR images are the input of the SR image reconstructs, and the output is a HR image.
In the following, we present two HR images:
    a- The first one respect the classic form of TV. (In chapter IV, section: IV.5.1)
    b- The second one respects the update form of TV.

Note: unfortunately, we have some problem with my memory laptop. In this reason we change the size of LR images input in SR reconstruct function (show appendix A) from 128x128 to 64x64 pixels. The result is:

A-using classic term of TV i.e. TV(X) =$\sum_i \sqrt{(\Delta_i^h X)^2 + (\Delta_i^v X)^2}$.

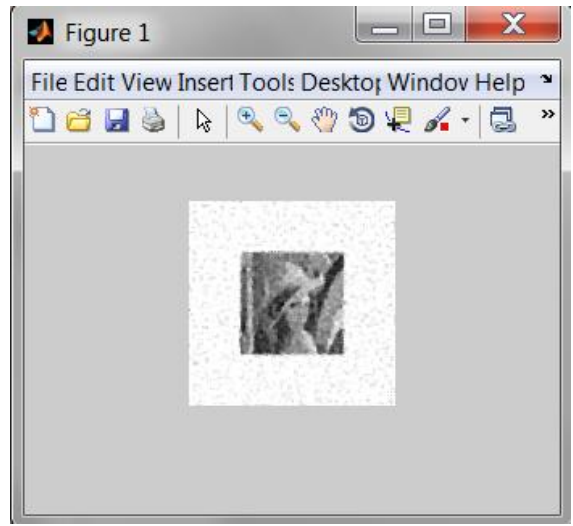

*Figure V.9 – HR image reconstruct use function deconvsuper appendix B.*

B-using update term of TV i.e. TV(X) = TV(X) = $\sum_i \sqrt{((\Delta_i^h X)^2 + (\Delta_i^v X)^2 + 1)}$ -1.

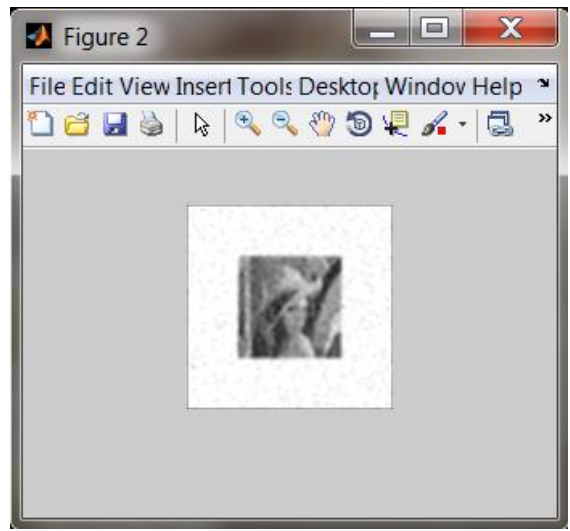

*Figure V.10 – HR image reconstruct use function deconvsuperplsone appendix B.*

The time need to compute this from LR images registration to HR image reconstruct is about 1 hour and half.

## V.5 - Conclusion

In this chapter, we use MATLAB simulator to compute the SR algorithm with classic and adaptive regulazer.

In fact we experiment three photos:

First one flower with colors and the second one lena have size 128x128. Here, we apply the SR algorithm with motionless.

Third image lena with white contour have size 128x128 but we reduce the size to 64x64 to compute the concept of motion.

The number of observation used N equal 8; relatively it is a small number give our algorithm a high performance.

# Chapter VI

# Conclusion and Future work

# Conclusion and Future work

We have presented a set of super-resolution algorithms, ranging from a more theoretical analysis of super resolution to a practical algorithm for low resolution. And we have developed a new majorization-minimization algorithm for image deconvolution under total variation regularization. The complexity of the algorithm is O(N) for finite support convolution kernels, where N is the number of image pixels.

Future Works

• Study of different compression techniques and their effects on the quality of reconstruction is not only essential for the optimal reconstruction of super-resolved images from compressed data, but also it is very important for designing novel compression techniques. A very interesting extension to our research is to focus on the design of a novel compression method which results in compressed low-quality images ideal for reconstruction by its matching super-resolution technique. Such method is of great importance for the design of efficient HD-TV video streams.

• Accurate subpixel motion estimation is an essential part of any image fusion process such as multi-frame super-resolution or demosaicing. The best extension for our algorithms is the incorporation of motion algorithms in the super resolution methods, and to build a convolution algorithm.

Really the work with images and obtain suitable result is a big joy but also we have out of memory problem. So we need a super machine to compute our model.
And the applied motion in our treatment is very important to generate the result from image to video sequence captured with a mobile phone.

SR image reconstruction is one of the most spotlighted research areas, because it can overcome the inherent resolution limitation of the imaging system and improve the performance of most digital image processing applications. I hope this work creates interest in this area as well as inspiration to develop and explicate the relevant techniques.

# Appendix A

```matlab
% Horn-Schunck optical flow method
% Horn, B.K.P., and Schunck, B.G., "Determining Optical Flow",
% Artificial  Intelligence Laboratory,Massachusetts Institute of
%Technology,Cambridge, MA 02139, U.S.A."
%
% Usage:
% [u, v] = HornSunck(im1, im2, alpha, ite)
% For an example, run this file from the menu Debug->Run or press (F9)
%
% -im1,im2 : two subsequent frames or images.
% -alpha : a parameter that reflects the influence of the smoothness term.
% -ite : number of iterations.
%
%
% Author: Mouhamd Chehaitly
% Lebanese University
% Ecole dictorale-Hadath
% Final Year Project 2011

% Default parameters

clc;
clear all;
close all;
I0=imread('car1.png');
I1=imread('car2.png');
im1=double(I0(:,:,1));
im2=double(I1(:,:,1));

[m n]=size(im1);
% Set initial o alpha and ite=iteration number
alpha=1;
ite=50;

% Set initial value for the flow vectors
uInitial = zeros(m,n);
vInitial = zeros(m,n);
u = uInitial;
v = vInitial;

% Estimate spatiotemporal derivatives
% Horn-Schunck original method

fx = conv2(im1,0.25* [-1 1; -1 1],'same') +…
    conv2(im2, 0.25*[-1 1; -1 1],'same');
fy = conv2(im1, 0.25*[-1 -1; 1 1], 'same') +…
    conv2(im2, 0.25*[-1 -1; 1 1], 'same');
ft = conv2(im1, 0.25*ones(2),'same') + conv2(im2, -0.25*ones(2),'same');

% Averaging kernel
kernel_1=[1/12 1/6 1/12;1/6 0 1/6;1/12 1/6 1/12];

% Iterations
```

```matlab
for i=1:ite
    % Compute local averages of the flow vectors
    uAvg=conv2(u,kernel_1,'same');
    vAvg=conv2(v,kernel_1,'same');
% Compute flow vectors constrained by its local average and the optical
flow constraints
u= uAvg - ( fx .* ( ( fx .* uAvg ) + ( fy .* vAvg ) + ft ) ) ./ ( alpha^2 +
fx.^2 + fy.^2);
v= vAvg - ( fy .* ( ( fx .* uAvg ) + ( fy .* vAvg ) + ft ) ) ./ ( alpha^2 +
fx.^2 + fy.^2);
end
%%
%Display Image

%plotFlow(u, v, im1, 2, 2);
figure
imshow(im1,[0 255]);
title(['optical flow : respect HS algorithm with alpha=1,…
        and iteration =',num2str(ite)]);
hold on;
rSize=2;
scale=5;
% Enhance the quiver plot visually by showing one vector per region
for i=1:size(u,1)
    for j=1:size(u,2)
        if floor(i/rSize)~=i/rSize || floor(j/rSize)~=j/rSize
            u(i,j)=0;
            v(i,j)=0;
        end
    end
end
quiver(abs(u),abs(v), scale, 'color', 'b', 'linewidth', 2);
set(gca,'YDir','reverse');

%other from to display image
figure;
v2 = 1:n;
v1 = 1:m;
[y,x] = meshgrid(v1,v2);
%[px,py] = gradient(double(I1),2,2);
imshow(uint8(im1)), hold on,quiver(v2,v1,u,v), hold off
```

# Appendix B

```matlab
%Abstrat - This is the iterative algorithm of using decent gradient
%described in the report to minimize the nonlinear equation using the total
%variation (TV) via MM algorithm as the regularization term. The output of
%this code will be the created super resolution image.
%Use HR images which are square where the size can be written 2^t !
% And the 2^t should be a multiple of the decimation factor.

% This program is created by:
% Mouhamad Chehaitly
% Lebanese University
% Ecole dictorale-Hadath
% Final Year Project 2011

% The software is provided "as is", without warranty of any kind.
clear all
close all
clc
start=tic;% start timer

%load the source image
hrimg = imread('images\Lena256.TIF');

% if the image is in RGB scale, so change it to gray scale.
if (size(hrimg,3) > 1)
  hrimg = rgb2gray(hrimg);
end

% hrimgscaled is the original image resized to 1/2 of the original
% size whith white edge.
% and create the warp or motion matrix.
[changeimage warpmatrix]=motionmatrix(hrimg,2);
hrimgscaled = changeimage;
figure ;
imshow(hrimgscaled);

% Determination of the size for the original image we consider in
imsize1= size(hrimgscaled,1); % nb of rows of the HR image
imsize2 = size(hrimgscaled,2); % nb of columns of the HR image

% transform image hrimg to vector(Lexicographic)
ImvectHR = double(reshape(hrimgscaled,imsize1*imsize2,1));

% i am use clear to refrech my ram and of course to accelerat the speed
clear hrimg  changeimage hrimgscaled;

% The candidate Blur Kernels are essentially the following :
ker1=(1/19) * [ 0 0 1 0 0 ; 0 1 2 1 0 ; 1 2 3 2 1 ; 0 1 2 1 0 ; 0 0 1 0 0];
ker2=(1/14) * [ 0 0 0 0 0 ; 0 1 2 1 0 ; 0 2 2 2 0 ; 0 1 2 1 0 ; 0 0 0 0 0];
ker3=(1/16) * [ 0 0 0 0 0 ; 0 1 2 1 0 ; 0 2 4 2 0 ; 0 1 2 1 0 ; 0 0 0 0 0];
ker4=(1/18) * [ 0 0 0 0 0 ; 0 1 2 1 0 ; 1 2 2 2 1 ; 0 1 2 1 0 ; 0 0 0 0 0];
ker5=(1/25) * [ 1 1 1 1 1 ; 1 1 1 1 1 ; 1 1 1 1 1 ; 1 1 1 1 1 ; 1 1 1 1 1];
ker6=(1/18) * [ 0 0 0 0 0 ; 0 2 2 2 0 ; 0 2 2 2 0 ; 0 2 2 2 0 ; 0 0 0 0 0];
ker7=(1/28) * [ 0 1 1 1 0 ; 1 1 2 1 1 ; 1 2 4 2 1 ; 1 1 2 1 1 ; 0 1 1 1 0];
ker8=(1/26) * [ 0 1 1 1 0 ; 1 1 2 1 1 ; 1 2 2 2 1 ; 1 1 2 1 1 ; 0 1 1 1 0];

% The decimation factor is decimfactor, it means that one line over
% decimfactor lines is considered as well for colums
```

```matlab
decimfactor = 2; % So the nb of rows for the observation is supposed
                 % imsize1/decimfactor and the nb of columns
                 % for the observation is imsize2/decimfactor
                 % One can imagine several decimation factors where each
                 % one is chosen for a particular transformation.
                 % But in our application, we suppose having the same
                 % decimation for all the observed images.

% Determination of the matrix of Blurred and Decimated, that multiply
% the HR image in order to
% obtain low resolution images (observations)
BD1 = Imagetransformation(imsize1,decimfactor,ker1);
BD2 = Imagetransformation(imsize1,decimfactor,ker2);
BD3 = Imagetransformation(imsize1,decimfactor,ker3);
BD4 = Imagetransformation(imsize1,decimfactor,ker4);
BD5 = Imagetransformation(imsize1,decimfactor,ker5);
BD6 = Imagetransformation(imsize1,decimfactor,ker6);
BD7 = Imagetransformation(imsize1,decimfactor,ker7);
BD8 = Imagetransformation(imsize1,decimfactor,ker8);

% Determination of the observation matrix:H
 H1 = BD1 * warpmatrix;
 H2 = BD2 * warpmatrix;
 H3 = BD3 * warpmatrix;
 H4 = BD4 * warpmatrix;
 H5 = BD5 * warpmatrix;
 H6 = BD6 * warpmatrix;
 H7 = BD7 * warpmatrix;
 H8 = BD8 * warpmatrix;

 clear BD1 BD2 BD3 BD4 BD5 BD6 BD7 BD8 warpmatrix;
% Blurring + decimation +warping transformation
ImageBlurred1 = H1 * ImvectHR;
ImageBlurred2 = H2 * ImvectHR;
ImageBlurred3 = H3 * ImvectHR;
ImageBlurred4 = H4 * ImvectHR;
ImageBlurred5 = H5 * ImvectHR;
ImageBlurred6 = H6 * ImvectHR;
ImageBlurred7 = H7 * ImvectHR;
ImageBlurred8 = H8 * ImvectHR;

% Estimation of the variance of the blurred Image VarImage Blurred;
VarImage1 = var(ImageBlurred1);
VarImage2 = var(ImageBlurred2);
VarImage3 = var(ImageBlurred3);
VarImage4 = var(ImageBlurred4);
VarImage5 = var(ImageBlurred5);
VarImage6 = var(ImageBlurred6);
VarImage7 = var(ImageBlurred7);
VarImage8 = var(ImageBlurred8);

% The 8 SNR values are for exemple (signal to noise ratio
% with the signal meaning the blurred image:
SNR_dB1 = 100;
SNR_dB2 = 50;
SNR_dB3 = 10;
SNR_dB4 = 20;
SNR_dB5 = 15;
SNR_dB6 = 30;
SNR_dB7 = 50;
```

```matlab
SNR_dB8 = 10;

% Estimation of the noise variance in terms of the SNR and the
% corresponding image variance for each of the 8 obesrvations
Varnoise1 = VarImage1 * 10^(-SNR_dB1/10);
Varnoise2 = VarImage2 * 10^(-SNR_dB2/10);
Varnoise3 = VarImage3 * 10^(-SNR_dB3/10);
Varnoise4 = VarImage4 * 10^(-SNR_dB4/10);
Varnoise5 = VarImage5 * 10^(-SNR_dB5/10);
Varnoise6 = VarImage6 * 10^(-SNR_dB6/10);
Varnoise7 = VarImage7 * 10^(-SNR_dB7/10);
Varnoise8 = VarImage8 * 10^(-SNR_dB8/10);

% Addition of noise to the blurred (and decimated) image.
ImvectLR1 = ImageBlurred1 + sqrt(Varnoise1) * randn(size(ImageBlurred1));
ImvectLR2 = ImageBlurred2 + sqrt(Varnoise2) * randn(size(ImageBlurred2));
ImvectLR3 = ImageBlurred3 + sqrt(Varnoise3) * randn(size(ImageBlurred3));
ImvectLR4 = ImageBlurred4 + sqrt(Varnoise4) * randn(size(ImageBlurred4));
ImvectLR5 = ImageBlurred5 + sqrt(Varnoise5) * randn(size(ImageBlurred5));
ImvectLR6 = ImageBlurred6 + sqrt(Varnoise6) * randn(size(ImageBlurred6));
ImvectLR7 = ImageBlurred7 + sqrt(Varnoise7) * randn(size(ImageBlurred7));
ImvectLR8 = ImageBlurred8 + sqrt(Varnoise8) * randn(size(ImageBlurred8));

clear SNR_dB1 SNR_dB2 SNR_dB3 SNR_dB4 SNR_dB5 SNR_dB6 SNR_dB7 SNR_dB8;
clear VarImage1 VarImage2 VarImage3 VarImage4 VarImage5 VarImage6 VarImage7 VarImage8;
clear Varnoise1 Varnoise2 Varnoise3 Varnoise4 Varnoise5 Varnoise6 Varnoise7 Varnoise8;

% Transformation from Lexico vector to Matrix in order to show the image
% result
ImtransformMatrix1 = reshape(ImvectLR1,imsize1/decimfactor,imsize2/decimfactor);
ImtransformMatrix2 = reshape(ImvectLR2,imsize1/decimfactor,imsize2/decimfactor);
ImtransformMatrix3 = reshape(ImvectLR3,imsize1/decimfactor,imsize2/decimfactor);
ImtransformMatrix4 = reshape(ImvectLR4,imsize1/decimfactor,imsize2/decimfactor);
ImtransformMatrix5 = reshape(ImvectLR5,imsize1/decimfactor,imsize2/decimfactor);
ImtransformMatrix6 = reshape(ImvectLR6,imsize1/decimfactor,imsize2/decimfactor);
ImtransformMatrix7 = reshape(ImvectLR7,imsize1/decimfactor,imsize2/decimfactor);
ImtransformMatrix8 = reshape(ImvectLR8,imsize1/decimfactor,imsize2/decimfactor);

% Show the transformed images or the LR images
figure ;
imshow(uint8(ImtransformMatrix1));
figure ;
imshow(uint8(ImtransformMatrix2));
figure ;
imshow(uint8(ImtransformMatrix3));
figure ;
imshow(uint8(ImtransformMatrix4));
figure ;
imshow(uint8(ImtransformMatrix5));
```

```matlab
figure ;
imshow(uint8(ImtransformMatrix6));
figure ;
imshow(uint8(ImtransformMatrix7));
figure ;
imshow(uint8(ImtransformMatrix8));

clear ImtransformMatrix1 ImtransformMatrix2 ImtransformMatrix3 ImtransformMatrix4;
clear ImtransformMatrix5 ImtransformMatrix6 ImtransformMatrix7 ImtransformMatrix8;
clear ImageBlurred1 ImageBlurred2 ImageBlurred3 ImageBlurred4;
clear ImageBlurred5 ImageBlurred6 ImageBlurred7 ImageBlurred8;

% Choice of the parameters for the super resolution operation,
% The choice is nearly at random !!
lamda =  1 ;
epsilon= 10^(-1);
iterations_num = 10;
tol = 0.01;

%change size of LR images from 128 to 32 because problem of mermory
Xest=deconvsuper(imresize(BD1,0.25),imresize(ImvectLR1,0.25),imresize(BD2,0.25),imresize(ImvectLR2,0.25),imresize(BD3,0.25),imresize(ImvectLR3,0.25),imresize(BD4,0.25),imresize(ImvectLR4,0.25),...
imresize(BD5,0.25),imresize(ImvectLR5,0.25),imresize(BD6,0.25),imresize(ImvectLR6,0.25),...
imresize(BD7,0.25),imresize(ImvectLR7,0.25),imresize(BD8,0.25),imresize(ImvectLR8,0.25),(imsize1/2)^2,lamda,epsilon,iterations_num, tol);

xest1=deconvsuperplsone(imresize(BD1,0.25),imresize(ImvectLR1,0.25),imresize(BD2,0.25),imresize(ImvectLR2,0.25),...

imresize(BD3,0.25),imresize(ImvectLR3,0.25),imresize(BD4,0.25),imresize(ImvectLR4,0.25),...

imresize(BD5,0.25),imresize(ImvectLR5,0.25),imresize(BD6,0.25),imresize(ImvectLR6,0.25),...

imresize(BD7,0.25),imresize(ImvectLR7,0.25),imresize(BD8,0.25),imresize(ImvectLR8,0.25),(imsize1/2)^2,lamda,epsilon,iterations_num, tol);

% Calculation of the MSE Mean Square error between original image and
% reconstructed one via method deconvsuper and deconvsuperplusone !
MSE = sum( (xest-ImvectHR)'* (xest-ImvectHR)) * (1/imsize1^2);
MSE1 = sum( (xest1-ImvectHR)'* (xest1-ImvectHR)) * (1/imsize1^2);

% Show the result of the Super resolution operation with both methods
Xresu1 = reshape(xest1,imsize1,imsize1);
figure;
imshow(uint8(Xresu1));
Xresu = reshape(xest,imsize1,imsize1);
figure;
imshow(uint8(Xresu));
telapsed = toc(start); %time needed to compute the registration LR images and reconstruct HR images.
```